%% file: main.tex
\documentclass[letterpaper]{article} 
\usepackage{aaai2027}
\usepackage[hyphens]{url} 
\usepackage{graphicx} 
\usepackage{natbib} 
\usepackage{caption} 
\usepackage{booktabs}
\usepackage{multirow}
\usepackage{amsmath}
\usepackage{amssymb}

\title{Do Current Retrievers Cover All the Evidence?\\
A Controlled Study of Conjunctive Cross-Page Retrieval}
\author{
Sungguk Cha, DongWook Kim, Mintae Kim,\\
Youngsub Han, Byoung-Ki Jeon, Sangyeob Lee
}
\affiliations{
LG Uplus, Seoul, South Korea\\
\{sungguk,dongwook92,iammt,yshan042,bkjeon,sangyeob\}@lguplus.co.kr
}
\nocopyright

\begin{document}
\maketitle

\begin{abstract}
Finding a long document relevant to a multi-part request is not the same as
establishing that it contains every requested piece of evidence. We study this
gap for conjunctive document retrieval, where two or three explicit conditions
must be supported on different pages of one document. We use n-Clue as a
controlled measurement instrument: 1{,}000 queries over 2{,}021 documents pair
all-condition golds with naturally occurring documents that satisfy only a
subset, and a complete-first success requires a top-10 gold to precede every
released subset qrel. Across 70 configurations, condition-wise decomposition
improves two dense backbones by 6.8--7.3 points and lexical--visual fusion adds
8.7, while four generic rerankers all reduce Gold-NDCG; these directions
replicate on a four-source stress set. Scaling one dense family from 0.6B to 8B
changes complete-first success by 0.0 points. The strongest displayed hybrid
illustrates the resulting gap: it finds a gold for 81.1\% of queries but
succeeds complete-first on only 35.8\%, and the gap persists across condition
count, target length, candidate density, query rendering, and the four-source
stress set. Finally, page-aware visual systems surface stored support for every
condition on only 5.1--5.3\% of queries. These results identify condition
coverage, rather than gold discovery alone, as the central bottleneck.
\end{abstract}

\section{Introduction}

Many consequential document searches are not requests for a topic but for a
conjunction of evidence. An analyst may need the report that contains an
executive summary, \emph{and} a grievance mechanism, \emph{and} a resettlement
policy. The three supporting passages may occur on pages 3, 24, and 51 of a
hundred-page report. Documents satisfying only one or two conditions are still
topically close---and may be more fluent or lexically similar matches---but they
cannot answer the request. A useful retriever must therefore do more than
\emph{discover} a relevant document: it must recognize condition coverage and
rank a document supported on every requested point ahead of natural partial
matches. This distinction is operationally important for retrieval-augmented
generation, because no reader or generator can cite evidence that retrieval
never supplies \citep{lewis2020rag}.

\begin{figure}[t]
\centering
\includegraphics[width=0.96\columnwidth]{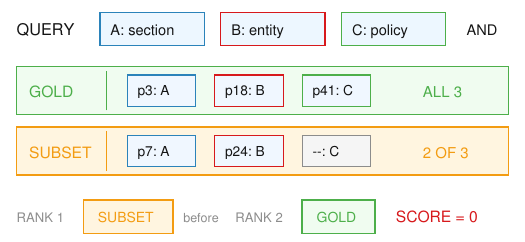}
\caption{Discovery and completion are different retrieval events. The gold
contains all three conditions on distinct pages; the subset contains only two.
Finding the gold at rank 2 is not complete-first success when the subset is
ranked first.}
\label{fig:concept}
\end{figure}

This requirement falls between established evaluation traditions. Broad IR
benchmarks reward topical relevance, and long-context benchmarks test whether a
model can retrieve or reason over distant content. Logical and multi-condition
benchmarks test compositional intent. Yet these traditions seldom jointly
control three facts that determine whether a long-document search is complete:
which conditions each corpus document satisfies, whether the gold evidence is
distributed across pages, and whether the most confusable alternatives are
real documents satisfying strict subsets of the request. Without those
controls, a high hit rate can mean that the system found a document on topic,
not that it distinguished full support from plausible incompleteness.

The distinction also exposes a modeling assumption. Joint dense retrieval,
page-level max pooling, and generic reranking ultimately compress a conjunction
and a long document into a scalar relevance score. Strong evidence for one
condition can then compensate for absent evidence for another. Increasing
candidate recall may even retrieve more complete documents without moving them
above highly relevant subsets. We call the resulting separation between
retrieving an all-condition document and ordering it before known partial
matches the \emph{discovery--completion gap}. Measuring this gap asks a more
diagnostic question than ``was a gold retrieved?'': once completion is
available in the top ranks, does the system recognize it?

We study that question using n-Clue as a controlled measurement instrument, not
as the endpoint of a benchmark-construction paper. CrossPage provides 1,000
queries over 2,021 real long documents. For each query, a document-wide
inventory records support for two or three explicit conditions and their page
locations; an all-condition gold must admit a distinct-page assignment, while
other corpus documents provide naturally occurring subset matches. The primary
event is intentionally strict and interpretable: a top-10 gold must precede
every released subset qrel (Figure~\ref{fig:concept}). We pair this event with
gold discovery and page-evidence diagnostics so that a low score can be
localized to candidate generation, complete-versus-subset ordering, or support
delivery.

Across 70 configurations, the gap is large and structurally informative.
Explicitly retrieving and combining condition-specific lists improves two dense
backbones by 6.8--7.3 points, and lexical--visual fusion adds 8.7 points, while
four tested generic rerankers all reduce Gold-NDCG; dense decomposition, fusion,
and reranker directions recur in a four-source document-level stress test, and
the main signs persist across query renderings. Scaling Qwen3-Embedding from
0.6B to 8B, by contrast, changes complete-first success by 0.0 points---a
bounded observation about one family, not a claim that scale can never help.
The strongest displayed hybrid illustrates the resulting gap: it retrieves a
gold for 81.1\% of queries but places a gold before the released subsets for
only 35.8\%, and page-aware visual systems surface stored support for every
condition on only 5.1--5.3\% of queries. The evidence therefore points away from
model size alone and toward explicit coverage state, complementary candidate
generation, and verification of the weakest condition.

Our contribution is empirical and diagnostic:
\begin{enumerate}
\item We formalize and measure the \textbf{discovery--completion gap}, separating
gold recall from the decision to rank complete evidence above natural subset
matches.
\item We use \textbf{controlled operation-level contrasts}---including
same-family scale, same-backbone decomposition, fixed-pool reranking, and
cross-modal fusion---to identify which system operations change that gap rather
than declaring a winner from a 70-run leaderboard.
\item We provide an \textbf{auditable page-grounded instrument} with complete
rankings, quote-backed label repair, task-factor and wording analyses, and
source-shift stress tests, enabling the claim and its boundaries to be checked.
\end{enumerate}
The intended conclusion is deliberately narrower than a claim about all
multi-evidence search: on explicit, conjunctive, cross-page requests, current
representative retrievers often find relevant long documents without
establishing that they cover the whole request.

\section{Related Work}

\paragraph{Retrieval diagnostics.}
MS~MARCO, BEIR, and MTEB established broad evaluation across retrieval domains
and representation families
\citep{nguyen2016msmarco,thakur2021beir,muennighoff2023mteb}. Their relevance
labels are appropriate for general search but do not normally expose which
parts of a conjunctive request each document supports. More recent diagnostics
move beyond topical match: BRIGHT emphasizes reasoning-intensive retrieval,
NevIR tests negation, and FollowIR and IFIR test instruction following
\citep{su2024bright,weller2024nevir,weller2025followir,song2025ifir}. We share
their goal of isolating a capability hidden by aggregate relevance, but target
a different failure event: an all-condition long document loses to a natural
document that is relevant to only a known subset.

\paragraph{Logical retrieval.}
MultiConIR directly evaluates multi-condition requests by comparing targets
against progressively weaker edited variants, while ComLQ covers conjunction,
disjunction, and negation and MCRank studies ranking under multiple, potentially
conflicting constraints
\citep{lu2025multiconir,xu2026comlq,pezeshkpour2025mcrank}. These studies make
clear that ordinary relevance models do not reliably preserve logical intent.
Our study complements them along the document axis: the retrieval unit is a
real long document, the required evidence occupies distinct pages, and the
partial competitors are independent documents already present in the corpus
rather than weakened edits of the target. This design permits separate
measurement of document discovery, complete-before-subset ordering, and
delivery of the pages supporting each condition.

\paragraph{Long-document evidence.}
LongEmbed evaluates embeddings over long textual contexts, whereas MMDocIR
centers multimodal page and layout retrieval
\citep{zhu2024longembed,dong2025mmdocir}. Counting-Stars tests access to multiple
evidence units within supplied long contexts, and FinMRAGBench and Doc-$V^*$
evaluate multi-page integration in financial RAG and interactive document VQA
\citep{song2025countingstars,yang2026finmragbench,zheng2026docvstar}. Multi-hop
QA similarly evaluates whether several facts can support an answer
\citep{yang2018hotpotqa,trivedi2022musique,ho2020twowiki}, while SciFact and
SciFact-Open attach evidence rationales to claim retrieval
\citep{wadden2020scifact,wadden2022scifactopen}. Those tasks primarily evaluate
page selection, context use, verification, or answer production once evidence
is available. Our estimand precedes those stages: among corpus documents,
does the ranker prefer the one containing the entire distributed evidence set
to documents missing a required part?

\paragraph{Ranking operations.}
The experimental matrix spans lexical and learned sparse retrieval
\citep{robertson2009bm25,formal2021splade}, dense encoders
\citep{karpukhin2020dpr,izacard2022contriever,wang2022e5,li2023gte,
lee2024nvembed,gunther2025jinaembeddingsv4,qwen3embedding2025}, late interaction
\citep{khattab2020colbert,santhanam2022colbertv2,chen2024bgem3}, and screenshot
retrieval \citep{faysse2024colpali,ma2024dse}. Standard hard-negative mining
finds confusable candidates for representation learning
\citep{xiong2021ance}; here, condition inventories instead identify why a
candidate is confusable and whether it is incomplete. We use condition-wise
aggregation \citep{fagin2003aggregation}, reciprocal-rank fusion
\citep{cormack2009rrf}, and generic reranking
\citep{nogueira2020monot5,sun2023rankgpt} as interventions. The purpose is not
to establish a universal ordering among architectures, but to determine whether
each operation improves discovery, coverage-sensitive ordering, both, or
neither.

\section{Study Design}

\subsection{Task and estimands}

A query $q=\{c_1,\ldots,c_n\}$ contains $n\in\{2,3\}$ explicit conditions.
Document $d$ has clue count $g(d,q)$ equal to the number of conditions supported
in the final document-wide evidence inventory. A \emph{gold} has $g=n$ and
admits a distinct supporting-page assignment; a \emph{subset} has
$1\leq g<n$. Let $r_G$ and $r_S$ be the best ranks of any gold and any released
positive subset qrel, with missing ranks equal to $\infty$. Our primary
complete-first estimand is
\begin{align}
\text{n-Clue Score@}k = \frac{100}{|Q|}\sum_{q\in Q}
\mathbb{1}[r_G\leq k\ \wedge\ r_G<r_S].
\end{align}
The score is a percentage, not a weighted composite. \textbf{Gold Hit@$k$}
measures discovery, and binary \textbf{Gold-NDCG@$k$} measures gold placement;
both use only $g=n$ as relevant and the same all-query denominator. Their
difference from n-Clue Score distinguishes failure to find a gold from failure
to place it before a known subset.

For page-scored systems, \textbf{Gold+Support@$k$} succeeds when the three
highest-scoring distinct surfaced pages of any top-$k$ gold collectively
intersect stored support for every condition. This is conservative: alternate
correct pages are not credited, but all queries remain in the denominator.
Systems receive query text, document text or page images, and corpus metadata,
never construction facts, qrels, or evidence-page annotations.

\paragraph{Why not graded relevance alone?}
Giving partial documents positive gain changes the estimand: it rewards a system
for ranking many subsets, although the task requires a complete document first
\citep{jarvelin2002ndcg}. On the raw matrix, BM25-OR exceeds BM25-AND on graded
NDCG@10 (.407 versus .265) but is worse at ordering gold over subsets
(pairwise AUC .586 versus .635). We therefore report partial documents as
diagnosed competitors rather than discounted successes. Gold Hit and n-Clue
Score are always paired: the former asks whether completion was available in the
candidate set, and the latter asks whether the system recognized it before a
known incomplete alternative.

\subsection{Controlled evidence testbed}

CrossPage contains 1{,}753 World Bank PDFs and 268 PMC OA articles. A
corpus-wide inventory of section and entity-phrase facts records their page IDs.
For every query, construction enumerates the union of documents containing at
least one condition and assigns its exact nonzero clue count. A gold must have a
nonempty page set for every condition, an empty all-condition page intersection,
and a distinct-page matching. Candidate conjunctions are rare even though their
conditions are individually common, yielding natural partial matches without
mining against a particular retriever.

\begin{table}[t]
\centering
\small
\setlength{\tabcolsep}{3.0pt}
\begin{tabular}{@{}lr@{}}
\toprule
CrossPage statistic & Value \\
\midrule
Queries (2 / 3 conditions) & 1,000 (814 / 186) \\
Documents / rendered pages & 2,021 / 41,441 \\
Pages per document (mean / median / max) & 20.5 / 11 / 446 \\
Shortest gold per query (mean / median pages) & 78.1 / 67 \\
Gold / subset qrels & 2,176 / 48,819 \\
Subset grade 1 / 2 & 46,795 / 2,024 \\
Distinct gold documents & 287 \\
Gold qrels (World Bank / PMC) & 2,175 / 1 \\
Single-page-solvable golds & 0 / 2,176 \\
Distinct-page-valid golds & 2,176 / 2,176 \\
\bottomrule
\end{tabular}
\caption{The controlled CrossPage instrument. Target opportunities are long
even though the full corpus includes short documents: for each query, the
shortest available gold has median length 67 pages.}
\label{tab:dataset}
\end{table}

An independent verifier re-derives every construction page set and both page
gates. Initial labels use the inventory plus an LLM read of fact-matched pages.
A blinded 300-pair audit then collects two exact-quote-backed judgments for 710
conditions. Exact grades agree on 254/300 pairs (84.7\%, weighted
$\kappa=.801$ \citep{cohen1960kappa}); disagreements receive expanded-page
adjudication. The final repair changes 20 CrossPage pairs (19 promotions, one
demotion), after which all 2{,}176 gold qrels still pass both page gates.
Reversing those observed corrections changes every clean-scorecard n-Clue Score
by at most 0.2 points and gives pre/post system-rank correlation $\rho=.9986$.
The supplement provides the construction genealogy, audit interfaces, transition
matrices, and sensitivity bounds.

Because CrossPage is World-Bank-centered, a separate \textbf{SourceShift} set
tests operation directions on arXiv, ERIC, Federal Register, and PMC OA. Its
final reviewed copy has 225 queries, 500 documents, and 8{,}841 nonzero qrels.
SourceShift is document-level and is not used to claim page-grounded
generalization.

\subsection{Systems and controlled interventions}

The study contains 70 CrossPage configurations: 16 standalone representations,
19 decomposition variants, 15 fusion variants, four rerankers, five LLM
paraphrase probes, and 11 baselines or validity controls. Fifty-four rows are
fair retrieval methods; 16 are controls or diagnostics. The supplement reports
every row, including unsuccessful variants.

\begin{table*}[t]
\centering
\small
\setlength{\tabcolsep}{3.5pt}
\begin{tabular}{p{0.17\textwidth}rp{0.29\textwidth}p{0.39\textwidth}}
\toprule
Matrix block & Runs & Controlled question & Capability readout \\
\midrule
Representation / scale & 16 & change encoder, size, interaction, or modality &
compare complete-first level to Gold Hit; isolate the 0.6B$\rightarrow$8B scale contrast \\
Condition composition & 19 & joint query $\rightarrow$ per-condition lists &
same-backbone min/product/harmonic aggregation and distinct-page BM25 \\
Complementarity & 15 & one list $\rightarrow$ equal-weight RRF &
test whether lexical, dense, and visual errors supply complementary golds \\
Fixed-pool reranking & 4 & candidate ranking $\rightarrow$ generic reranker &
ask whether stronger general relevance also rewards full condition coverage \\
LLM wording probes & 5 & raw query $\rightarrow$ frozen paraphrase &
test whether low dense performance is only a list-rendering artifact \\
Validity controls & 11 & mask conditions, expose facts, or restrict pages &
check condition necessity, leakage ceiling, and cross-page floor \\
\midrule
Submitted matrix & 70 & 54 fair methods + 16 controls/diagnostics &
every row and uncertainty interval appears in Supplementary Table S1 \\
\bottomrule
\end{tabular}
\caption{Capability-study matrix. The additional robustness program reruns 13
systems under three deterministic forms (39 rankings), recomputes task-factor
slices from final qrels, and applies the same operations to SourceShift.}
\label{tab:program}
\end{table*}

Standalone systems span BM25, BGE-M3 sparse/multi-vector, E5, GTE, Jina-v4,
Qwen3-Embedding 0.6B/4B/8B, ColBERTv2, ColQwen2.5, Qwen3-VL 2B/8B, and
DSE-Qwen2. Condition-wise decomposition retrieves every released condition
separately and aggregates its lists. Its distinct-page form is
\begin{equation}
S_{\mathrm{SDR}}(d,q)=
\max_{\pi\ \mathrm{injective}}\min_i s(c_i,p_{\pi(i)}),
\end{equation}
where $\pi$ assigns different pages to conditions. It uses no labels or stored
evidence pages. Visual page aggregation averages the three best page scores.
Fusion is equal-weight RRF ($k=60$, depth 100). Rerankers operate on fixed
candidate pools. Models are frozen inference-only, and labels tune neither
weights nor fusion parameters.

The primary scorecard evaluates 13 representative systems on the released clean
semicolon query. Controlled intervention effects come from the original
raw-form matrix because those endpoints were fixed there. For blocks with
several variants, the displayed endpoint is the highest raw-form n-Clue Score;
all alternatives remain reported. An additional 39-run matrix reruns 13 systems
under raw, clean, and deterministic natural renderings. Five dense systems also
have cache-backed LLM paraphrases. These probes vary wording, not the qrels.

\subsection{Inference and dependence}

CrossPage intervals resample 453 exact-gold-set query families with 1{,}000
bootstrap draws. Paired randomization flips complete families. For each of eight
controlled blocks, max-$|T|$ first accounts for all tried endpoints, followed by
Holm correction across blocks. The gold-membership graph links 995/1{,}000
queries, so the supplement additionally reports crossed query/document
weighting, leave-one-document analyses, common-support ordering, fixed-denominator
failure decompositions, and judged-subset-promotion bounds
\citep{efron1979bootstrap}. SourceShift comparisons are stress analyses on its
final target-reviewed qrels.

\section{Results}

\subsection{RQ1: Discovery is not complete-first ranking}

\begin{table*}[t]
\centering
\small
\setlength{\tabcolsep}{1.5pt}
\input{tables/table_primary_scorecard.tex}
\caption{Official clean-query CrossPage scorecard. \textbf{n-Clue Score@10}
requires a top-10 all-condition gold to precede every released subset. Gold
Hit@10 measures discovery, and Gold-NDCG@10 measures gold placement. All rows
use only released inputs. $\dagger$ marks the hybrid endpoint selected post hoc
on the separate raw-form matrix.}
\label{tab:scorecard}
\end{table*}

Table~\ref{tab:scorecard} shows a large discovery--completion gap for every
approach. The selected lexical--visual hybrid has the highest displayed values:
81.1 Gold Hit but only 35.8 n-Clue Score, a 45.3-point gap. ColQwen top-3 finds a
gold on 71.2\% of queries but completes the event on 25.2\%. Thus its page
aggregation mainly expands discovery; it does not establish that every
condition is covered before partial matches. Even systems with lower absolute
recall exhibit 30-point gaps. A gold-only metric or hit rate therefore
overstates readiness for multi-evidence retrieval.

\begin{table}[t]
\centering
\small
\setlength{\tabcolsep}{1.0pt}
\input{tables/table_failure_decomposition.tex}
\caption{Mutually exclusive clean-query outcomes (\% of queries). ``Subset
first'' means a gold is in the top 10 but a released subset precedes it; the
three columns sum to 100.}
\label{tab:failure}
\end{table}

The outcome decomposition makes the bottleneck sharper
(Table~\ref{tab:failure}). The hybrid reduces outright discovery misses to
18.9\%, yet subset-first ordering remains its largest outcome at 45.3\%.
ColQwen top-3 behaves similarly: aggregation reduces misses to 28.8\%, while
46.0\% of queries contain a retrieved gold that still follows a subset.
Condition-wise Qwen retrieval lowers misses relative to the joint 8B system, but
35.5\% of queries become subset-first failures. Better coverage can therefore
move mass from ``miss'' to ``retrieved but incomplete-first'' without solving
the ranking decision.

One released example illustrates the event. Query
\texttt{nclue\_0000} requests an Executive Summary, COVID-19, and Sanctions. Its
gold stores support on pages 2, 19, and 60. Clean BM25 ranks a grade-1 subset
first and the gold fourth; visual top-3 ranks subsets first and ninth but misses
the gold; their RRF hybrid moves the gold to rank 2, still behind a grade-1
subset at rank 1. The hybrid has found the right 69-page document, but the
ranking still provides no complete-first guarantee. The supplement shows the
released pages for this gold and a grade-2 subset.

\begin{figure*}[t]
\centering
\includegraphics[width=0.97\textwidth]{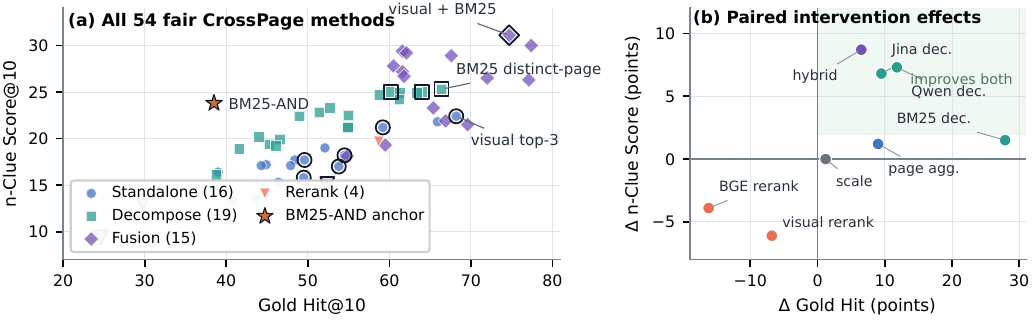}
\caption{Raw-form diagnostic matrix. (a) All 54 fair methods under Gold Hit@10
and n-Clue Score@10. (b) Controlled variant-minus-base effects. The upper-right
quadrant improves both discovery and completion; other quadrants expose why
recall gains alone are insufficient.}
\label{fig:landscape}
\end{figure*}

\subsection{RQ2--RQ3: Scale is insufficient; explicit coverage helps}

Figure~\ref{fig:landscape} shows the full fair-method landscape.
Table~\ref{tab:contrasts} then changes one operation at a time. Effect columns
are percentage points; the paired score difference is the decision quantity.

\begin{table*}[t]
\centering
\small
\setlength{\tabcolsep}{2.5pt}
\begin{tabular}{p{0.12\textwidth}p{0.25\textwidth}rcrrr}
\toprule
Question & Controlled comparison (base $\rightarrow$ variant) & Score &
$\Delta$Score [paired 95\% CI] & $\Delta$Hit & $\Delta$NDCG &
Sel.+Holm $p$ \\
\midrule
Scale & Qwen3-Emb. 0.6B $\rightarrow$ 8B & 17.7$\rightarrow$17.7 & 0.0 [$-2.7$,2.8] & +1.2 & +0.4 & 1.000 \\
Page aggregation & ColQwen2.5 $\rightarrow$ top-3 pages & 21.2$\rightarrow$22.4 & +1.2 [$-1.6$,4.0] & +9.0 & +3.3 & 1.000 \\
Lexical composition & BM25-AND $\rightarrow$ distinct-page dec. & 23.8$\rightarrow$25.3 & +1.5 [$-3.1$,5.9] & +27.9 & +9.8 & 1.000 \\
Dense composition & Jina joint $\rightarrow$ harmonic dec. & 18.2$\rightarrow$25.0 & +6.8 [3.7,9.9] & +9.5 & +7.9 & .005 \\
Dense composition & Qwen 0.6B joint $\rightarrow$ product dec. & 17.7$\rightarrow$25.0 & +7.3 [3.9,10.6] & +11.8 & +8.0 & .002 \\
Fusion & visual top-3 $\rightarrow$ visual+BM25 RRF & 22.4$\rightarrow$31.1 & +8.7 [5.6,11.9] & +6.5 & +8.9 & .005 \\
Text reranking & BGE-M3 $\rightarrow$ BGE reranker & 13.3$\rightarrow$9.4 & $-3.9$ [$-7.1$,$-0.7$] & $-16.2$ & $-5.9$ & .168 \\
Visual reranking & ColQwen2.5 $\rightarrow$ Qwen-VL 8B & 21.2$\rightarrow$15.1 & $-6.1$ [$-9.6$,$-2.8$] & $-6.8$ & $-6.7$ & .008 \\
\bottomrule
\end{tabular}
\caption{Controlled CrossPage contrasts. Intervals resample exact-gold-set
families. Each $p$-value accounts for every tried endpoint in its block and then
Holm correction across the eight blocks.}
\label{tab:contrasts}
\end{table*}

\paragraph{Model size is not the missing operation.}
Within the tested Qwen3-Embedding family, increasing size from 0.6B to 8B adds
1.2 Gold-Hit points but 0.0 n-Clue Score points. This is a bounded conclusion
about one family, not a claim that scale can never help. Across the standalone
encoders tested here, however, no representation reaches the composition or
fusion endpoints.

\paragraph{Decomposition helps two dense backbones.}
Separating conditions improves n-Clue Score by 6.8 points for Jina and 7.3 for
Qwen, with selection-aware corrected $p=.005$ and $.002$. The operation, not a
larger checkpoint, supplies the gain. Yet composition is not sufficient:
distinct-page BM25 raises Gold Hit by 27.9 points but Score by only 1.5,
retrieving many more golds without reliably moving them above subsets.

\paragraph{Complementarity helps; generic relevance reranking does not.}
The selected lexical--visual RRF endpoint improves both discovery and completion
(+8.7 Score, corrected $p=.005$), consistent with complementary lexical and
visual evidence. All four tested rerankers lower Gold-NDCG relative to their
fixed-pool bases. The displayed visual reranker lowers Score by 6.1 points and
survives correction; the displayed text decrease does not. A generic relevance
objective therefore cannot be assumed to reward the weakest uncovered
condition.

\subsection{RQ4: The gap persists across task and distribution shifts}

\begin{table*}[t]
\centering
\small
\setlength{\tabcolsep}{2.7pt}
\input{tables/table_capability_strata.tex}
\caption{Descriptive clean-query slices. Cells are \textbf{n-Clue Score / Gold
Hit}. $L_{\min}$ is the shortest gold-document length because any gold can
satisfy a query; $|D^+|$ is the nonzero-qrel pool. Slices overlap and are not
causal contrasts.}
\label{tab:strata}
\end{table*}

The discovery--completion gap is not confined to one easy slice
(Table~\ref{tab:strata}). For the hybrid, moving from two to three conditions
raises Gold Hit from 80.5 to 83.9 but lowers Score from 37.0 to 30.6, expanding
the gap from 43.5 to 53.2 points. In dense positive pools, it still finds a gold
on 81.8\% of queries but scores 31.0. On queries whose shortest gold exceeds 100
pages, the hybrid remains 46.2 points below its Hit, and the visual top-3 system
is 46.1 points below. These descriptive slices show that discovery alone remains
insufficient as the conjunction and distractor set grow.

The wording probes reach the same qualitative conclusion. Across five dense
retrievers, moving from raw construction text to an LLM paraphrase changes
n-Clue Score by only $-1.0$ to $+0.6$ points; all five remain between 12.8 and
18.5. These paraphrases are not guaranteed lexical-free, so they are a semantic
wording probe rather than a paraphrase benchmark. In the separate
13-system$\times$3-form matrix, dense composition improves under raw, clean, and
deterministic natural renderings; the BGE and visual rerankers decrease under
all three. Absolute BM25 levels do change, so we do not pool forms into one
leaderboard. Condition-dropping adds a necessity check: removing either of the
first two conditions lowers Gold-NDCG for all five dense systems; removing the
third on $n=3$ lowers it for three of five.

\begin{table}[t]
\centering
\small
\setlength{\tabcolsep}{3.0pt}
\begin{tabular}{lrrc}
\toprule
Operation (variant $-$ base) & CrossPage & SourceShift & Direction \\
\midrule
Page aggregation & +1.2 & $-0.9$ & flip \\
Lexical composition & +1.5 & $-20.0$ & flip \\
Jina composition & +6.8 & +6.7 & same \\
Visual + lexical RRF & +8.7 & +15.1 & same \\
Text reranking & $-3.9$ & $-8.0$ & same \\
Visual reranking & $-6.1$ & $-3.1$ & same \\
\bottomrule
\end{tabular}
\caption{Raw-form operation-direction stress test on the final reviewed qrels.
Values are variant-minus-base n-Clue Score@10 points.}
\label{tab:source-shift}
\end{table}

Under source shift, dense composition, cross-modal complementarity, and the two
displayed reranker directions replicate (Table~\ref{tab:source-shift}).
Lexical distinct-page composition reverses sharply, and page aggregation also
flips. The study therefore supports dense decomposition and complementarity as
the more stable directions, not a source-general claim for every task-aware
operation.

\subsection{RQ5: Retrieved golds rarely arrive with all stored support}

\begin{table}[t]
\centering
\small
\setlength{\tabcolsep}{3.0pt}
\begin{tabular}{lrrr}
\toprule
Page-scored run & Cond. coverage & Full coverage & Gold+Support \\
\midrule
Qwen3-VL 2B & 49.5 & 17.2 & 5.3 \\
Qwen3-VL 8B & 47.4 & 13.6 & 5.1 \\
\bottomrule
\end{tabular}
\caption{Stored-page evidence coverage (\%). Condition and full coverage are
conditional on a top-10 gold in the page-scored evidence run; Gold+Support uses
all 1,000 queries.}
\label{tab:evidence}
\end{table}

For the two visual systems that retain page scores, Gold+Support@10 is 5.3 for
Qwen3-VL 2B and 5.1 for 8B. Conditional on first retrieving a gold, a stored
reference page is surfaced for an individual condition about half the time, but
all-condition page coverage is only 14--17\% (Table~\ref{tab:evidence}). For 2B,
full coverage falls from 21.2\% on two-condition hits to 5.1\% on
three-condition hits; the corresponding 8B values are 14.1\% and 11.3\%. The
metric credits only stored pages and is therefore conservative. Even so, the
order-of-magnitude drop from Gold Hit to evidence delivery identifies the next
modeling target: condition-to-page matching or iterative retrieve-and-verify
behavior that explicitly rewards the weakest condition.

\section{Construct Validity}

We bracket the intended construct with controls rather than infer it from the
headline gap alone. A privileged construction-fact counter reaches 100 n-Clue
Score, whereas requiring all conditions on one page scores zero; independently,
0/2,176 golds are single-page-solvable and all 2,176 admit a distinct-page
assignment. The task is therefore executable with the defining evidence facts
but cannot be solved from a single page. Conditions are also not decorative:
dropping $c_1$ or $c_2$ lowers Gold-NDCG by 1.9--6.5 points for each of five
dense retrievers, and $c_3$ lowers three of five significantly.

Wording and shortcut controls delimit alternative explanations. Five frozen
LLM paraphrases change dense scores by only $-1.0$ to $+0.6$ points, and the
main operation signs mostly persist across 39 deterministic-form runs. A
length-matched gold--subset analysis changes distinct-page BM25 AUC from .606
to .598, while a query-independent condition-frequency prior reaches only 14.1
Score. Finally, reversing the 20 observed CrossPage audit repairs moves the
clean scorecard by at most 0.2 points. Supplementary Table S5 gives the full
masking rows, intervals, and adversarial promotion bounds.

These controls do not turn the instrument into a population sample. The
paraphrases retain lexical overlap, the audit is stratified rather than
exhaustive, and length matching addresses only one observed confound. Our
conclusions therefore concern the fixed, versioned instrument and controlled
operation contrasts; they do not estimate the prevalence of multi-evidence
needs in real traffic.

\section{Implications for Retrieval Design}

The controlled results separate three system responsibilities that are often
collapsed into one relevance score. \emph{Discovery} places at least one gold in
the candidate set. Page aggregation, decomposition, and fusion can all improve
this stage. \emph{Coverage tracking} records which query conditions have support
in each candidate and where that support occurs. The same-backbone decomposition
gains show that retaining condition-specific scores is useful, but BM25's
$+27.9$ Hit / $+1.5$ Score contrast shows that simply unioning more condition lists is
not enough. Finally, \emph{verification and ordering} must reject a topically
strong candidate when its weakest condition is unsupported. The reranking and
Gold+Support results locate the largest remaining gap at this stage.

A direct design follows from this separation. Instead of compressing a
conjunction and a long document into one scalar immediately, a retriever can
maintain a condition--page matrix
$M_{i,p}=s(c_i,p)$. Candidate generation may remain lexical, dense, visual, or
fused. A coverage layer then finds a distinct-page assignment and exposes both
the assigned evidence and the weakest condition score. A verifier reads the
small assigned page set, updates condition states, and either ranks the document
as complete or retrieves again for the missing condition. This design combines
the observed benefit of decomposition with an explicit safeguard against the
subset flooding seen in Table~\ref{tab:failure}.

The study also suggests a training signal. Natural grade-$(n-1)$ documents are
targeted counterexamples: highly relevant to most of the query yet required to
lose to a complete document. A condition-aware reranker should therefore
optimize a complete-over-subset contrast and return evidence assignments rather
than generic pointwise relevance alone. This is a hypothesis for model
development, not a method result; the released rankings and page labels make it
directly testable.

\section{Scope, Limitations, and Release}

The study supports a narrow claim about \emph{explicit conjunctive, cross-page
document retrieval}. Conditions are named sections or phrases, not arbitrary
implicit information needs. CrossPage is World-Bank-centered, and its gold
judgments overrepresent long documents; the target-length slices describe this
structure but do not make it representative of all corpora. SourceShift broadens
document sources but lacks page evidence. The released scorer treats
inventory-zero pairs as grade 0, so complete-first success is relative to a
versioned evidence inventory rather than exhaustive new judgments of all two
million query--document pairs. Sampled quote-backed repair and promotion
sensitivity reduce, but cannot eliminate, false-negative risk.

The strongest hybrid was selected post hoc on the raw matrix and is descriptive.
The roster is broad but does not include every proprietary retriever, an
end-to-end long-context reader, or a trained retrieve-and-verify agent.
Consequently, we diagnose a retrieval bottleneck rather than claiming an upper
bound on complete RAG systems. Future work should test semantic paraphrases with
strict lexical controls, add page evidence to multiple sources, and compare
task-aware readers or agents against the fixed retrieval outputs.

\paragraph{Reproducibility.}
Code, annotations, and frozen rankings are released at
\url{https://github.com/sunggukcha/n_clue}: final queries, qrels, all 70
rankings, the 39 form rankings, scoring and analysis code, audit records, model
revisions, hashes, and prompts. One command, using only the Python standard
library and no network access, regenerates the principal score, contrast, and
sensitivity outputs and verifies them against frozen claims; a second script
regenerates the failure and task-factor tables. Source URLs and recorded
terms support reconstruction without
redistributing third-party PDFs, images, or OCR \citep{gebru2021datasheets}.

\section{Conclusion}

Representative retrievers often find a relevant long document without
establishing that it covers the full request. On the controlled CrossPage
instrument, the interventions that change this are explicit condition-wise
decomposition and lexical--visual fusion, which improve two dense backbones and
the hybrid; generic reranking is frequently harmful, and scaling one dense
family does not help. The strongest displayed hybrid illustrates the gap these
operations only partly close: 81.1 Gold Hit but only 35.8 complete-first
success, and page-aware systems deliver all stored support on about 5\% of
queries. Retrieval must therefore track each condition and verify its evidence
before declaring a document complete.

\clearpage
\bibliography{references}

\end{document}


\maketitle
\tableofcontents
\clearpage

\section{Study Questions and Measurement Instrument}

The paper is a capability study, with n-Clue serving as the controlled
measurement instrument. The distinction matters for reading the supplement:
construction and audit sections establish that the instrument measures the
intended complete-before-subset event, while the experiment sections establish
what current retrieval operations do on that event. The study questions and
their supporting artifacts are:

\begin{center}
\small
\setlength{\tabcolsep}{4pt}
\begin{tabular}{p{0.10\textwidth}p{0.47\textwidth}p{0.31\textwidth}}
\hline
Question & Capability being tested & Principal evidence \\
\hline
RQ1 & Can a retriever place an all-condition document before natural
condition-subset matches? & clean scorecard; complete matrix \\
RQ2 & Does representation choice or within-family scale close the
discovery--completion gap? & 16 standalone rows; Qwen 0.6B/4B/8B contrast \\
RQ3 & Which operations improve condition coverage? & 19 decomposition, 15
fusion, and four fixed-pool reranking rows \\
RQ4 & Do conclusions survive task-factor, wording, and source shifts? &
Table S1j; Tables S2--S4; SourceShift \\
RQ5 & Do page-aware systems deliver stored support for every condition? &
Gold+Support and conditional page coverage \\
\hline
\end{tabular}
\end{center}

\subsection{Construction and validation genealogy}

\begin{figure*}[t]
\centering
\includegraphics[width=0.98\textwidth]{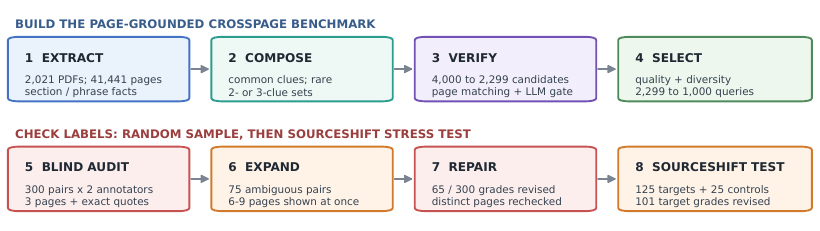}
\caption{The measurement-instrument pipeline. The top row turns a corpus-wide
page-fact inventory into 1,000 controlled CrossPage queries. The bottom row
tests structural validity, repairs quote-backed sampled labels, and separately
stress-tests high-impact SourceShift labels.}
\label{fig:construction-pipeline}
\end{figure*}

CrossPage pays the page-analysis cost once and reuses a versioned, document-wide
condition inventory. For a condition $c_i$, $P_0(d,c_i)$ is the set of pages in
document $d$ whose stored facts establish that condition. Construction
enumerates every document in $\bigcup_i\{d:P_0(d,c_i)\ne\varnothing\}$ and emits
its exact nonzero clue count. A gold requires every page set to be nonempty,
their all-condition intersection to be empty, and a distinct-page assignment
to exist. A subset has at least one but fewer than all conditions. The table
below records which immutable or repaired copy supports each reported result.

\begin{center}
\scriptsize
\setlength{\tabcolsep}{3.2pt}
\begin{tabular}{p{0.12\textwidth}p{0.19\textwidth}p{0.35\textwidth}p{0.23\textwidth}}
\hline
Stage / copy & Input $\rightarrow$ output & Purpose & Reported use \\
\hline
Mine $P_0$ & 2,021 PDFs / 41,441 pages $\rightarrow$ condition--page sets &
analyze pages once and retain page provenance & immutable construction facts \\
Compose $P_0$ & 4,000 $\rightarrow$ 2,299 verified $\rightarrow$ 1,000 selected
queries & rare conjunction, semantic gate, distinct-page matching, fixed
quality/diversity selection & 50,995 nonzero CrossPage pairs \\
Blind audit & 300 pairs / 710 conditions $\rightarrow$ two votes & test the
inventory under hidden IDs with exact quotes & 84.7\% exact-grade agreement \\
Repair $P$ & 75 expanded pairs / 94 conditions $\rightarrow$ final page sets &
pool six--nine pages and rerun clue count plus both page gates & 2,176 gold +
48,819 subsets; all CrossPage results \\
Gate $S_0$ & eight sources $\rightarrow$ four passing sources & apply the frozen
98\% extraction gate and document-level conjunction labels & 225 queries /
500 documents after sampled repair \\
Stress $S$ & 125 targets + 25 controls $\rightarrow$ blind tie-breaks & challenge
every reviewed label able to reduce the frozen fusion--BM25 contrast & final
8,841-qrel SourceShift copy \\
\hline
\end{tabular}
\par\smallskip\textbf{Measurement genealogy.} $P_0\!\rightarrow\!P$ is the
page-grounded CrossPage path; $S_0\!\rightarrow\!S$ is the separate
document-level SourceShift stress path. Query rendering changes text, never
qrels.
\end{center}

The final CrossPage inventory has 2,176 gold and 48,819 subset qrels. An
independent re-derivation finds zero single-page-solvable golds and a valid
distinct-page assignment for every gold. The initial sampled audit, expanded
adjudication, actual interfaces, and exact transition matrices are reported in
the integrated Evidence Audit and Adjudication section below.

\section{Capability Matrix and Main-Paper Derivation}

The capability study is a contrast matrix rather than a single-model leaderboard.
Its 70 rows vary representation, composition, page aggregation, fusion partner,
reranking, and query rendering while preserving the same 1,000 query IDs and
qrels. The categories and row counts below are read directly from the submitted
reference-run manifest; no unsuccessful run is silently removed.

\begin{center}
\small
\begin{tabular}{lrl}
\hline
Matrix block & Runs & Experimental role \\
\hline
Standalone retrievers & 16 & lexical/sparse/dense/late-interaction/visual \\
Decomposition & 19 & five backbones $\times$ min/product/harmonic; BM25 SDR \\
Fusion & 15 & lexical, dense, and visual complementarity \\
Reranking & 4 & text and visual pools; 2B/8B controls \\
Query-form controls & 5 & cached rendering robustness \\
Baselines/diagnostics & 11 & lexical/frequency controls, leakage oracle, floor \\
\hline
Total & 70 & all have 1,000 public-ID ranking rows \\
\hline
\end{tabular}
\end{center}

\noindent\textbf{Execution settings.}
The table below collects the shared settings behind the matrix; exact model IDs
and cached revisions are included in
\path{provenance/model_revisions.json}.

\begin{center}
\small
\setlength{\tabcolsep}{4pt}
\begin{tabular}{p{0.20\textwidth}p{0.68\textwidth}}
\hline
Component & Frozen setting \\
\hline
Corpus representation & MinerU page OCR; dense text chunks use 3,072 tokens with 256-token overlap \\
Retrieval depth & top 100 documents per retriever and fusion; all reported metrics at rank 10 \\
Composition & separate condition lists with min, product, harmonic mean, or injective distinct-page aggregation \\
Visual aggregation & page MaxSim; displayed page-pooling variant averages the three highest page scores \\
Fusion / reranking & equal-weight RRF ($k=60$, depth 100); text reranking uses a fixed top-100 pool and visual reranking a fixed top-20 pool \\
Randomness / ties & selection seed 11; 1,000 bootstrap replicates; 10,000 randomization draws; stable input order resolves score ties \\
Model identity & frozen inference-only checkpoints; exact repository IDs and revisions in the archive provenance file \\
\hline
\end{tabular}
\end{center}

Main-paper Table~1 summarizes the controlled instrument, Table~2 maps the
70-run capability study, Table~3 is the clean-query scorecard, and Table~4
decomposes clean-query outcomes: n-Clue Score is simply 100 times the fraction
of all queries with a top-10 gold before every judged subset. Gold Hit and
Gold-NDCG are supporting subscores. Figure~2 exposes all 54 fair methods, and
Table~5 reports eight controlled raw-form base--variant pairs:
Qwen scale; ColQwen page aggregation; BM25/Jina/Qwen explicit composition;
cross-modal fusion; and text/visual reranking. Each pair changes one named
operation and reports the base and variant together. The table is therefore a
contrast summary, not a mini-leaderboard. For blocks with multiple aggregation
variants, however, the displayed variant is the highest raw-form n-Clue Score
endpoint and is explicitly post hoc; the complete table below prevents hiding
the remaining variants. Blockwise max-$|T|$ tests include every tried endpoint,
then use Holm correction across the eight displayed blocks, so inference reflects
selection across the complete block. Main-paper Table~6 and the Table~4 failure
decomposition are regenerated from the final clean rankings and qrels by
\path{make_capability_strata.py}; Table~7 compares operation directions under
source shift. Main-paper Tables~8--9 report page-evidence delivery and construct
validity, respectively.
Figure~2(a) retains all 54 fair method rows, and Figure~2(b) places paired
effects in the Gold-Hit--n-Clue-Score delta plane. Table~S1 below is the
complete result: Tables~S1a--S1c rank the 66 rows that do not read construction
facts in three readable blocks, while Table~S1d isolates the four privileged fact diagnostics without a
system rank. The leading fair block is dominated by fusion and decomposition,
whereas all four generic rerankers fall below their own candidate-pool bases.
\input{tables/table_s1.tex}

\section{Headline-Contrast Sensitivity}

The next three tables apply the same checks to all eight main-paper pairs. They
separate ordering on a common success support from unconditional failure modes,
cross query/document dependence with every tried endpoint represented, and
adversarial promotion of currently judged high-ranked subsets. The last check is
an exact bound over judged-subset promotions; it does not cover inventory-zero
pairs, which follow the released grade-0 scoring rule.

\input{tables/table_headline_sensitivity.tex}

\section{Clustered Uncertainty and Paired Tests}

The code/data supplement includes the public-release-native outputs:
\begin{itemize}
\item \texttt{../data\_release/reference\_runs/clustered\_ci.md}
\item \texttt{../data\_release/reference\_runs/clustered\_ci.json}
\item \texttt{../data\_release/reference\_runs/paired\_tests.md}
\item \texttt{../data\_release/reference\_runs/paired\_tests.json}
\item \texttt{../data\_release/reference\_runs/headline\_sensitivity.json}
\item \texttt{paper/v2\_revalidation\_summary.json}
\end{itemize}
These files are computed from
\path{queries.jsonl}, \path{qrels.jsonl}, \path{corpus.jsonl}, and
\path{reference_runs/*.jsonl}. The family bootstrap groups queries with the
same complete set of gold document IDs (453 families in CrossPage); it addresses exact
duplicate construction families. The paired-test file covers decomposition,
fusion, and reranker contrasts. For the eight main-paper rows it also reports
paired exact-family bootstrap intervals. The headline-sensitivity file adds
blockwise max-$|T|$ family sign flips over every tried endpoint and Holm
correction across eight blocks. CrossPage tests remain exploratory. The frozen SourceShift
fusion--BM25 contrast is confirmatory with respect to system choice, but its qrel
sensitivity is itself an evaluated result.

Table~S1i extends the family analysis to nonidentical queries sharing a document
by crossing independent query and document weights, first over gold
memberships and then over every judged gold/partial membership. These are stress
models, not uniquely correct sampling distributions.

\begin{center}
\small
\begin{tabular}{lrr}
\hline
Shared-document sensitivity & CrossPage & SourceShift \\
\hline
Queries / distinct gold docs & 1{,}000 / 287 & 225 / 218 \\
Connected components (largest) & 5 (995) & 23 (195) \\
Observed fusion$-$BM25 gap & +.1718 & +.0161 \\
Crossed query/doc 95\% interval & [.1352, .2063] & [-.0364, .0743] \\
Leave-one-gold-doc gap range & [.1620, .1856] & [.0097, .0278] \\
Remove 20 highest-degree docs & +.0420 (53 q.) & +.0437 (121 q.) \\
All-judged components (largest) & 1 (1{,}000) & 1 (225) \\
All-judged crossed 95\% interval & [.1537, .1899] & [-.0210, .0550] \\
All-judged leave-one-doc range & [.1420, .1920] & [-.0024, .0681] \\
Remove 20 highest-degree judged docs & +.0997 (153 q.) & +.0920 (89 q.) \\
\hline
\end{tabular}
\end{center}

\noindent\textbf{Table S1i.} Shared-document dependence for the separately frozen
fusion--BM25 comparison. The gold graphs are
near-connected in CrossPage; including subsets makes both graphs fully connected.
CrossPage remains positive, whereas both SourceShift crossed intervals include zero after
targeted revalidation. Deletion sharply reduces the available query set. The
complete JSON and generator are in the archive.

\section{Task-Factor Slices (Table S1j)}

Main-paper Table~6 is reproduced below. It is computed directly from final
CrossPage qrels and the five clean-query rankings, rather than inherited from
the pre-repair metrics cache.

\begin{center}
\scriptsize
\setlength{\tabcolsep}{2.7pt}
\input{tables/table_capability_strata.tex}
\end{center}

\noindent\textbf{Table S1j.} Descriptive task-factor slices. Every cell is
n-Clue Score@10 / Gold Hit@10 in percentage points. $L_{\min}$ is the minimum
page count among a query's gold documents because the scored event can succeed
with any gold. $|D^+|$ is the number of nonzero inventory qrels. The slices
overlap and are not causal contrasts.

The table makes three boundaries explicit. First, an extra condition is not
merely an extra retrieval anchor: for the hybrid, Gold Hit rises from 80.5 to
83.9 between $n=2$ and $n=3$, while complete-first success falls from 37.0 to
30.6. The 0.6B decomposed Qwen row is more extreme (61.1/28.3 to 61.3/14.0).
Second, target length affects paradigms differently. BM25 drops from 38.4 to
24.1 Score between the short and $>100$-page slices, whereas Qwen-8B does not;
the discovery--completion gap nevertheless remains at least 27.8 points for
every displayed system on the long slice. Third, positive-inventory density is
consistently adverse to complete-first ranking: every displayed system has lower
Score for $|D^+|>50$ than for $|D^+|\leq25$, even when its Gold Hit changes
little. These are descriptive observations because condition frequency, source,
length, and pool size are correlated by construction.

\path{capability_strata.json} stores all unrounded cells, counts, definitions,
and the expected aggregate-score assertions.
\path{make_capability_strata.py} regenerates both the JSON and LaTeX table and
fails if any aggregate clean-scorecard value drifts.

\section{Query-Form Robustness Protocol}

The public release separates the measurement definition from the query rendering.
The query IDs, qrels, corpus IDs, gold documents, and partial documents are
frozen. Only the display/query text varies across forms:
\begin{itemize}
\item \texttt{raw}: exact \texttt{query\_text\_eval\_full} used in the original
70-system matrix.
\item \texttt{clean\_v2}: each cleaned concept is quoted and separated by a
semicolon, e.g., \texttt{"Grievance Mechanism"; "Comparison"; "Resettlement"}.
\item \texttt{natural}: deterministic template
\texttt{Find the document that discusses "X", "Y", and "Z".}
\end{itemize}
The semicolon delimiter is deliberately chosen because 388/1{,}000 older public
queries contained concept-internal \texttt{and}, making clue boundaries visually
ambiguous. The file \path{../data_release/release/queries_forms.jsonl}
contains all required rows for the server-side form-ablation matrix. The QA
report \path{../data_release/release/query_forms_report.md} checks
verbatim concept containment and reports 1{,}000/1{,}000 coverage for
\texttt{raw}, \texttt{clean\_v2}, and \texttt{natural}. LLM paraphrases are
cache-backed only; they are not invented when no paraphrase cache is present.
All three public forms are deterministic renderings of the same structured
conditions. They isolate sensitivity to delimiters and list-like wording while
holding the information need fixed.

\subsection{LLM paraphrase probe (Table S2a)}

The original 70-run matrix also contains five cache-backed \texttt{glm-5.2}
paraphrase runs for dense retrievers. Unlike the deterministic natural template,
these may substitute concepts (for example, ``executive overview'' for
``executive summary''). They were generated once and frozen before evaluation.

\begin{center}
\small
\begin{tabular}{lrrr}
\hline
Dense retriever & Raw Score & Paraphrase Score & $\Delta$ \\
\hline
BGE-M3 & 13.3 & 12.8 & $-0.5$ \\
Jina-v4 & 18.2 & 18.5 & +0.3 \\
Qwen3-Embedding 0.6B & 17.7 & 16.7 & $-1.0$ \\
E5-large & 17.2 & 17.8 & +0.6 \\
GTE-large & 13.5 & 13.3 & $-0.2$ \\
\hline
\end{tabular}
\end{center}

\noindent\textbf{Table S2a.} Raw-to-LLM-paraphrase n-Clue Score@10. The narrow
$[-1.0,+0.6]$ range shows that the low dense-retriever level is not explained by
the raw list syntax alone. This is a probe, not a lexical-free paraphrase
benchmark: several generated queries retain condition words, and only five
dense systems were run.

\section{Query-Form Robustness Results (Table S2b)}

The server-side S1 run produced 39 ranking files:
13 systems/intermediates $\times$ 3 forms. The reported matrix has 13 rows
(\texttt{bge\_m3\_joint} is the shared intermediate base for text rerankers)
over \texttt{raw}, \texttt{clean\_v2}, and \texttt{natural}. Three planned
freshness anchors---Jina v5 text, Jina v5 omni, and ColQwen3---are explicitly
documented as blocked in
\texttt{../data\_release/reference\_runs/forms/BLOCKED\_SYSTEMS.md}; they are
not silent omissions. One freshness anchor, \texttt{rerank\_qwen3r8b}, was
runnable because its weights were already cached.

\begin{center}
\scriptsize
\begin{tabular}{lccc}
\hline
System & Raw & Cleaned & Natural \\
\hline
\texttt{bm25\_and} & 23.8 & 26.8 & 26.8 \\
\texttt{bm25\_decompose\_sdr} & 25.3 & 27.9 & 27.9 \\
\texttt{text\_jinav4} & 18.2 & 17.2 & 18.6 \\
\texttt{text\_qwen3emb8b} & 17.7 & 19.6 & 19.5 \\
\texttt{jinav4\_dec\_hmean} & 25.0 & 24.1 & 24.1 \\
\texttt{qwen3emb0.6b\_dec\_product} & 25.0 & 25.6 & 25.6 \\
\texttt{colqwen2.5} & 21.2 & 22.3 & 24.0 \\
\texttt{colqwen2.5\_ptop3} & 22.4 & 25.2 & 24.4 \\
\texttt{bm25and\_colqwen2.5ptop3} & 31.1 & 35.8 & 35.6 \\
\texttt{bge\_m3\_joint} & 13.3 & 12.4 & 12.7 \\
\texttt{rerank\_bge\_m3} & 9.4 & 9.6 & 10.9 \\
\texttt{rerank\_qwen3vl8b} & 15.1 & 17.3 & 16.7 \\
\texttt{rerank\_qwen3r8b} & 13.2 & 15.0 & 12.1 \\
\hline
\end{tabular}
\end{center}

Values are n-Clue Score@10 recomputed from the public-ID form
rankings and adjudicated qrels. The form matrix is used for controlled
operation comparisons rather than as a second leaderboard: fusion and dense
composition improve the score across all three renderings; the BGE and visual
rerankers lower it across all three, while the Qwen3 text reranker is mixed. The
Qwen3 decomposition row compares a
0.6B decomposed run against the stronger 8B joint anchor because the server
matrix reran only the 8B joint Qwen3 retriever; the full 70-system paired test
also includes the same-backbone 0.6B joint/decomposed comparison.

\noindent\textbf{Table S2b.} Complete 13-system deterministic query-form matrix.
These 39 runs, unlike Table S2a, use only verbatim concept-preserving renderings.

\section{SourceShift Robustness Study (Tables S3--S4)}

SourceShift construction began with 1{,}000 PDFs balanced across eight sources
(125 each). Although global MinerU extraction passed at 62{,}039/62{,}838 pages
(98.73\%), four sources failed the pre-specified per-source 98\% gate:
FDA/DailyMed (95.47\%), NASA NTRS (95.37\%), USGS (96.01\%), and World Bank
(96.08\%). The final quality-gated evaluation removes those sources' queries, documents,
qrels, and ranking entries, initially retaining 226 queries. The stratified audit
then demoted the only gold for one ERIC query; that zero-gold query and its nine
qrels are removed, producing the 225-query, 8{,}849-qrel post-audit copy $S_0$.
The 125-target challenge described below removes eight
additional zero-grade rows and promotes reviewed targets. The resulting qrel
copy $S$ has 500 documents, 225 queries, 8{,}841 nonzero qrels, 502 gold judgments,
and 218 distinct gold documents. Query counts remain arXiv 42, ERIC 44, Federal
Register 125, and PMC OA 14; 198 have $n{=}2$ and 27 have $n{=}3$.
The 125-target challenge was selected on the frozen raw fusion--BM25
gold-only NDCG@10 contrast. Tables S3--S4 instead apply n-Clue Score uniformly to
the final qrels, so their operation comparisons are a separate estimand.

\begin{center}
\scriptsize
\begin{tabular}{lccc}
\hline
System & Raw & Cleaned & Natural \\
\hline
\texttt{bm25\_and} & 44.4 & 32.4 & 32.4 \\
\texttt{bm25\_decompose\_sdr} & 24.4 & 17.8 & 17.8 \\
\texttt{text\_jinav4} & 16.0 & 15.6 & 17.3 \\
\texttt{text\_qwen3emb8b} & 16.9 & 14.7 & 16.9 \\
\texttt{jinav4\_dec\_hmean} & 22.7 & 20.9 & 20.9 \\
\texttt{qwen3emb0.6b\_dec\_product} & 23.1 & 21.3 & 21.3 \\
\texttt{colqwen2.5} & 15.6 & 16.0 & 20.0 \\
\texttt{colqwen2.5\_ptop3} & 14.7 & 16.0 & 19.1 \\
\texttt{bm25and\_colqwen2.5ptop3} & 29.8 & 27.1 & 28.4 \\
\texttt{bge\_m3\_joint} & 15.6 & 10.7 & 10.2 \\
\texttt{rerank\_bge\_m3} & 7.6 & 7.1 & 8.0 \\
\texttt{rerank\_qwen3vl8b} & 12.4 & 12.9 & 15.6 \\
\texttt{rerank\_qwen3r8b} & 12.0 & 9.8 & 12.4 \\
\hline
\end{tabular}
\end{center}

\noindent\textbf{Table S3.} n-Clue Score@10 for all 13
SourceShift systems and query forms after the targeted qrel challenge.

\begin{center}
\scriptsize
\begin{tabular}{lrrrrrr}
\hline
& \multicolumn{2}{c}{Raw} & \multicolumn{2}{c}{Cleaned} & \multicolumn{2}{c}{Natural} \\
Contrast & $\Delta$ & $p$ & $\Delta$ & $p$ & $\Delta$ & $p$ \\
\hline
Fusion $-$ BM25 & $-14.7$ & .0001 & $-5.3$ & .1579 & $-4.0$ & .2984 \\
Fusion $-$ visual top-3 & +15.1 & .0001 & +11.1 & .0005 & +9.3 & .0018 \\
Lexical decomp. $-$ BM25 & $-20.0$ & .0001 & $-14.7$ & .0001 & $-14.7$ & .0001 \\
Jina decomp. $-$ joint & +6.7 & .0364 & +5.3 & .0745 & +3.6 & .2623 \\
Qwen decomp. $-$ joint & +6.2 & .0604 & +6.7 & .0340 & +4.4 & .2145 \\
BGE reranker $-$ joint & $-8.0$ & .0038 & $-3.6$ & .2120 & $-2.2$ & .4158 \\
Qwen-VL reranker $-$ base & $-3.1$ & .3760 & $-3.1$ & .3245 & $-4.4$ & .1693 \\
Qwen3 reranker $-$ base & $-3.6$ & .2682 & $-0.9$ & .8543 & +2.2 & .4638 \\
\hline
\end{tabular}
\end{center}

\noindent\textbf{Table S4.} Query-paired n-Clue Score differences in points with
two-sided 10{,}000-sample permutation tests ($n=225$). Rows are exploratory.
On this target-reviewed four-source copy, adding lexical evidence to the visual
list keeps its positive direction, as do dense composition and the BGE/visual
reranker signs; lexical decomposition changes direction and the Qwen3 text
reranker is mixed.

For the 13 raw runs, graded-NDCG and gold-only-NDCG rankings have Spearman
$\rho=.5714$ after revalidation. The largest movement is eight ranks:
\texttt{bm25\_and} moves from twelfth under graded NDCG to fourth under gold-only
NDCG. Raw fusion-minus-BM25 is heterogeneous by source: +0.1283 arXiv, +0.0587
ERIC, -0.0428 Federal Register, and +0.0718 PMC OA.

SourceShift is intentionally a document-level conjunctive robustness study;
CrossPage remains the page-grounded benchmark. Its 477 condition records have no
stored evidence pages, so Tables S3--S4 compare document-ranking operation
directions on the final target-reviewed four-source qrel copy. The released audit
records and final qrels make that scope executable.

\section{Construct Diagnostics and Label Examples (Tables S5a--S6)}

The main scorecard is intentionally small enough to read as a capability result.
Tables S5a--S5b retain the secondary experiments that test whether its
interpretation survives condition removal, metric choice, query construction,
page grounding, and sampled label repair. They are diagnostic experiments, not
additional competing systems.

\begin{center}
\small
\setlength{\tabcolsep}{4pt}
\begin{tabular}{lrrr}
\hline
Dense retriever & Drop $c_1$ & Drop $c_2$ & Drop $c_3$ ($n=3$) \\
\hline
Jina-v4 & $-4.6^*$ & $-6.5^*$ & $-5.1^*$ \\
Qwen3-Embedding 0.6B & $-3.9^*$ & $-5.0^*$ & $-6.0^*$ \\
BGE-M3 & $-1.9^*$ & $-4.6^*$ & $-3.1^*$ \\
E5-large & $-4.7^*$ & $-5.2^*$ & $-2.5$ \\
GTE-large & $-1.9^*$ & $-3.2^*$ & $-2.9$ \\
\hline
\end{tabular}
\end{center}

\noindent\textbf{Table S5a.} Gold-NDCG@10 change in points after dropping one
condition. $c_1/c_2$ use all 1,000 queries; $c_3$ uses the 186 three-condition
queries. $^*$ denotes a paired bootstrap interval excluding zero
($p\leq.05$). Dropping $c_1$ or $c_2$ also lowers Gold Hit by 6.8--14.0 points
for every model. Thus every query position carries retrieval signal; the
complete-first failure is not generated by one decorative condition.

\begin{center}
\scriptsize
\setlength{\tabcolsep}{3pt}
\begin{tabular}{p{0.18\textwidth}p{0.17\textwidth}p{0.28\textwidth}p{0.27\textwidth}}
\hline
Diagnostic & Scope & Main observation & Supported interpretation \\
\hline
Metric ranking & 13 SourceShift raw runs & graded vs. gold-only $\rho=.571$; maximum shift 8 ranks & partial-credit metrics do not measure released complete-first success \\
Query rendering & 13 systems $\times$ 3 forms & BM25 level changes; dense composition and reranking directions persist & use the clean scorecard for levels and paired tests for operations \\
Condition masking & 5 dense systems; each position & masking positions 1/2 lowers NDCG by .020--.065; position 3 is mixed & positions 1--2 are consistently load-bearing across the five systems \\
Construction controls & fact oracle + single-page floor & oracle $\approx1.0$; single-page AND $=0$ & fact access is leakage; stored gold evidence is cross-page \\
Difficulty covariates & rarity, $n$, length, pool size & $\beta_n=+.03$ ($p=.27$); rarity $-.41$, pool size $+.34$ & more conditions are not automatically harder after controlling density \\
Page grounding & two visual runs with page scores & conditional reference hit $\approx50\%$; Gold+Support .051--.053 & stored-support delivery is a separate modeling target \\
Observed audit repairs & 20 CrossPage corrections & 19 subset$\rightarrow$gold; 1 gold$\rightarrow$subset; 12 appear in clean top-10 union & on submitted clean rankings: Score change $\leq$.2 points; rank $\rho=.9986$ \\
Promotion stress & all 8 pairs; 0--3 judged promotions/query & every positive gap can cross zero at $k{=}1$ & prioritize pooled judgments in benchmark maintenance \\
\hline
\end{tabular}
\end{center}

\noindent\textbf{Table S5b.} Construct-validity and secondary diagnostic
experiments. For each top-10 gold, Gold+Support takes its three highest-scoring
distinct surfaced pages; it succeeds if any retrieved gold's page set intersects
stored support for every condition. Thus several golds use the best-covered one,
and the surfaced pages need not form an injective assignment. It uses all 1,000
queries as the denominator, and stable ranking order resolves score ties.
Per-condition and 14--17\%
all-condition page-coverage rates are conditional on first retrieving a gold.
The complete per-system pre/post values and construction-union assertions are in
\path{crosspage_audit_impact.json} and are regenerated by
\path{analyze_crosspage_audit_impact.py}.

The difficulty regression jointly includes condition rarity, condition count,
gold-document length, and judged-pool size. On a density-matched subset,
three-condition items are easier for every tested system because an additional
condition can also provide an extra retrieval anchor. Matching document length
changes AUC only slightly (e.g., distinct-page BM25 .606$\rightarrow$.598).
These checks rule out simple condition-count and document-length accounts after
controlling the reported covariates.

\noindent\textbf{Released query example.}
\texttt{nclue\_0000} asks for
\texttt{"Executive Summary"; "COVID-19"; "Sanctions"}. Table S6 follows that
same query across one gold and two naturally occurring subset documents. Page
IDs are the released rendered-page IDs; a dash means that the released
document-wide evidence inventory contains no page for that condition.

\begin{center}
\scriptsize
\setlength{\tabcolsep}{3pt}
\begin{tabular}{p{0.10\textwidth}p{0.34\textwidth}p{0.15\textwidth}p{0.15\textwidth}p{0.15\textwidth}}
\hline
Label & Target (release ID) & Executive Summary & COVID-19 & Sanctions \\
\hline
Gold $g=3$ &
\texttt{world\_bank\_34454344} &
$\checkmark$ p.2 &
$\checkmark$ p.19 &
$\checkmark$ p.60 \\
Subset $g=2$ &
\texttt{world\_bank\_34422285} &
--- &
$\checkmark$ p.28 &
$\checkmark$ p.47 \\
Subset $g=1$ &
\texttt{world\_bank\_34353633} &
--- &
--- &
$\checkmark$ p.4 \\
\hline
\end{tabular}
\end{center}

\noindent\textbf{Table S6.} Concrete gold and subset labels for one released
three-condition query. The gold has evidence for all three conditions on
distinct pages. The grade-2 and grade-1 targets retain two and one condition,
respectively, so the scorer requires a system to place the gold first.

\begin{figure}[ht]
\centering
\small
\setlength{\fboxsep}{1pt}
\textbf{Gold target: all three conditions on distinct pages}\\[-2pt]
\begin{minipage}[t]{0.315\textwidth}
\centering
\fbox{\includegraphics[width=0.96\linewidth]{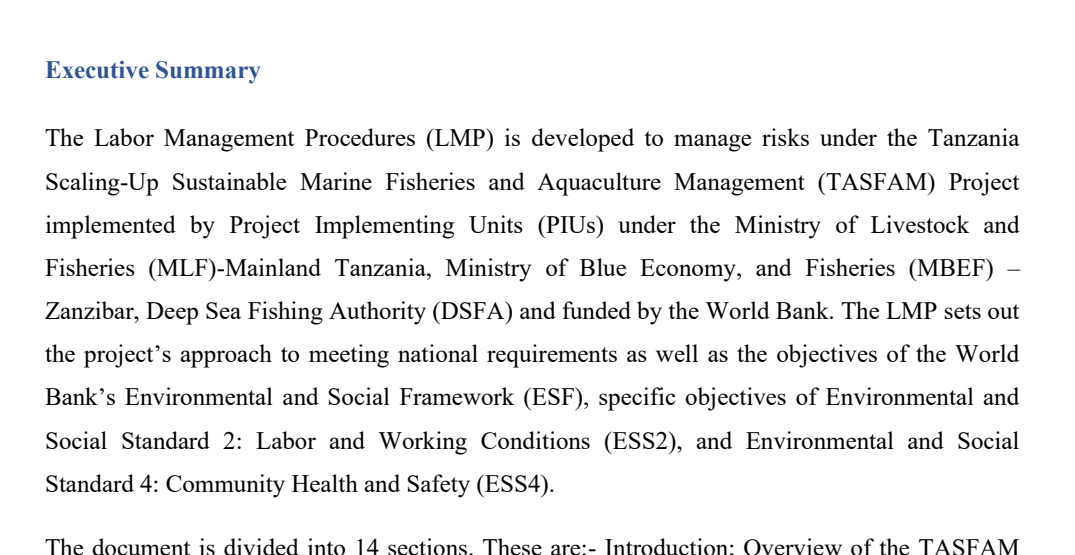}}\\[-1pt]
\textbf{$\checkmark$ Executive Summary, p.2}
\end{minipage}\hfill
\begin{minipage}[t]{0.315\textwidth}
\centering
\fbox{\includegraphics[width=0.96\linewidth]{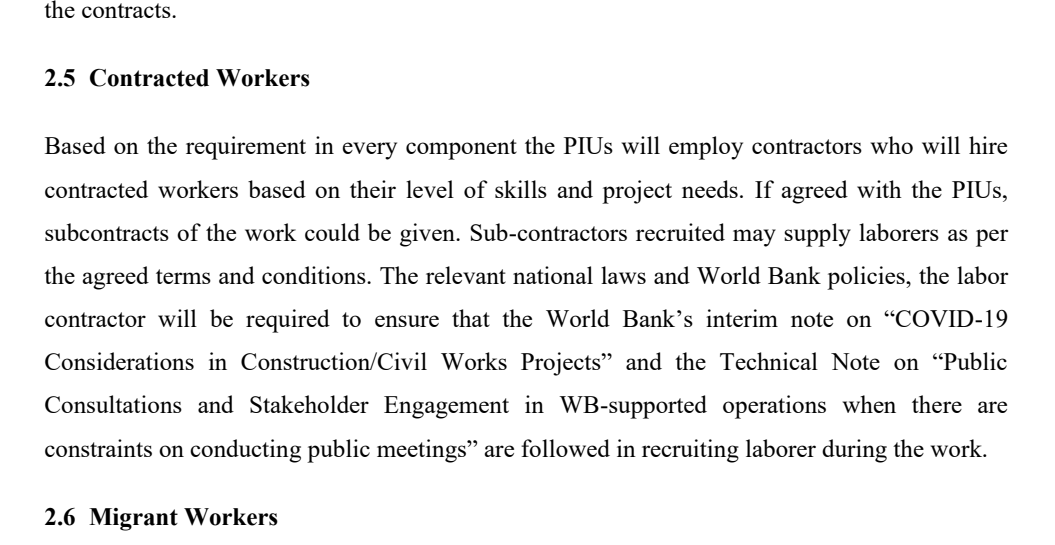}}\\[-1pt]
\textbf{$\checkmark$ COVID-19, p.19}
\end{minipage}\hfill
\begin{minipage}[t]{0.315\textwidth}
\centering
\fbox{\includegraphics[width=0.96\linewidth]{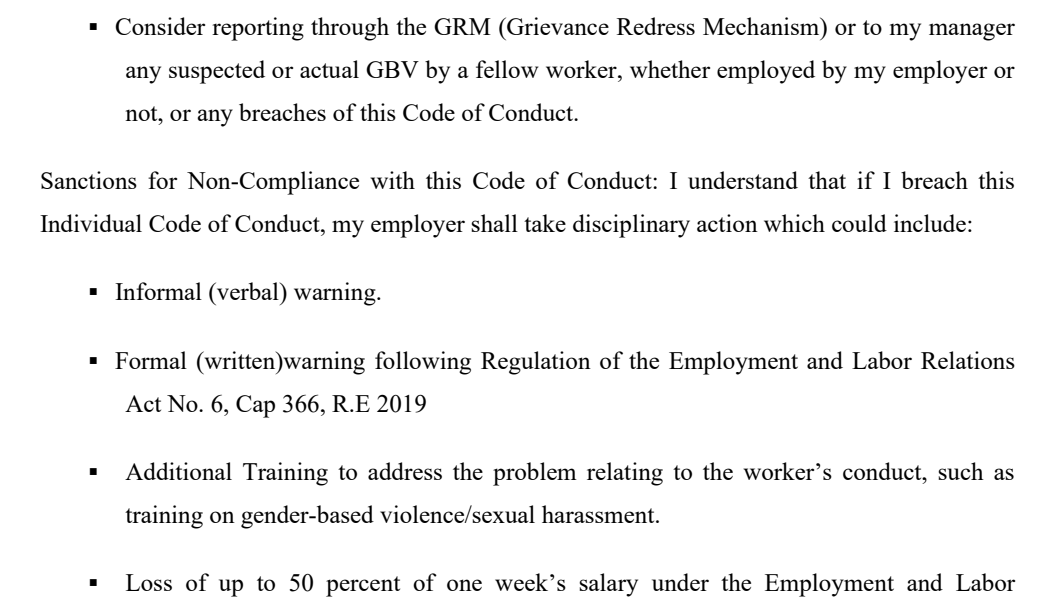}}\\[-1pt]
\textbf{$\checkmark$ Sanctions, p.60}
\end{minipage}

\vspace{7pt}
\textbf{Subset target: one condition absent, so grade 2 of 3}\\[-2pt]
\begin{minipage}[t]{0.315\textwidth}
\centering
\fbox{\parbox[c][0.105\textheight][c]{0.92\linewidth}{\centering
\textbf{No released page}\\[3pt]
for Executive Summary}}\\[-1pt]
\textbf{--- Executive Summary}
\end{minipage}\hfill
\begin{minipage}[t]{0.315\textwidth}
\centering
\fbox{\includegraphics[width=0.96\linewidth]{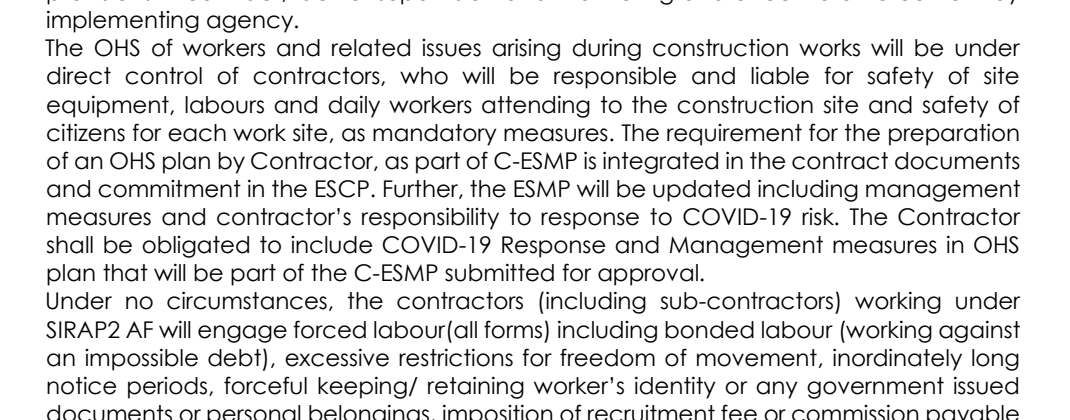}}\\[-1pt]
\textbf{$\checkmark$ COVID-19, p.28}
\end{minipage}\hfill
\begin{minipage}[t]{0.315\textwidth}
\centering
\fbox{\includegraphics[width=0.96\linewidth]{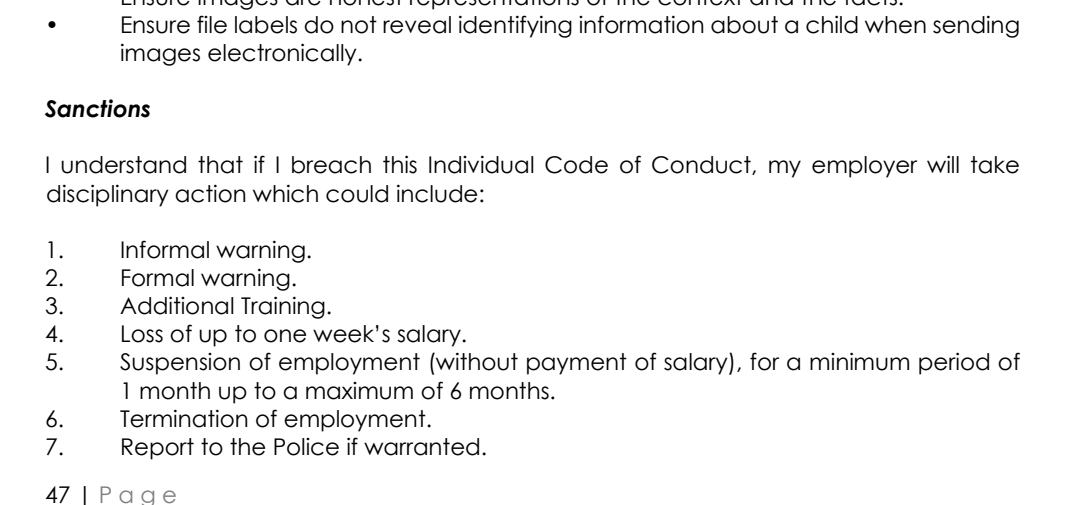}}\\[-1pt]
\textbf{$\checkmark$ Sanctions, p.47}
\end{minipage}
\caption{Actual target-page excerpts for the gold and grade-2 subset in Table
S6. The shared query is identical; the label changes because the gold covers
all three conditions on separate pages whereas the subset has no stored
Executive Summary page.}
\label{fig:label-example}
\end{figure}

\setcounter{table}{6}
\input{supplement_audit.tex}

\section{Construction-Model Provenance}

CrossPage semantic clustering and item verification use \texttt{glm-5.2} through the
same Anthropic-compatible endpoint. The exact prompts and response parsers are
included in \path{judge_clusters.py} and \path{verify_items.py}. Both disable
extended thinking, specify a maximum output length, and do not set a temperature,
top-$p$, or provider seed; therefore the provider default sampling policy applies.
Cluster calls return KEEP/SPLIT/REJECT JSON, with a second validated partition
call only for SPLIT. Item calls return one Boolean and evidence string per
condition after reading the pages where mined surface forms occur. Any mismatch
between this semantic read and the mined condition set removes the complete item;
there is no pre-audit majority vote or human--LLM conflict-resolution stage.
Prompts, accepted outputs, source revision, and final qrels are frozen, enabling
artifact reproduction even though fresh API calls are not bit-reproducible.

For each query, construction enumerates the union of the per-condition document
sets and emits every member with its exact nonzero clue count. Across CrossPage,
the sum of those union sizes is 50{,}995, exactly the number of qrel rows. The
other 1{,}970{,}005 query--document pairs have clue count zero under the same
inventory; their zero rows are omitted from the compact JSONL and restored by the
scorer. Equal-score ties retain stable input order.

\section{Reconstruction and Copyright}

The released benchmark data contain annotations, query text, qrels, document
metadata, reference rankings, and scripts. They deliberately exclude source PDFs,
rendered page images, OCR text, and page crops. This is a copyright and license
decision, not a storage-capacity decision: the local OCR dump is small enough for
Hugging Face, but it is extracted third-party full text from a mixed-license
corpus.

Each CrossPage corpus row records its source, landing/direct-PDF URL, SHA-256, and the
terms known at collection time. PMC OA rows retain article-level licenses;
World Bank rows are explicitly marked for per-document terms verification. All
rights remain with source owners, and the release asserts no new ownership over
source documents or short evidence quotations. The included
\path{RELEASE_AND_REMOVAL_POLICY.md} provides an ID-based correction/removal
procedure through the documented contact channel and commits the public
release to a direct contact and change manifest.

CrossPage content reconstruction is script-based. All 2{,}021 compact CrossPage
corpus rows retain their original-source \texttt{pdf\_url} and SHA-256. A fresh
network-enabled check on July 22, 2026 downloaded a deterministic sample of ten
World Bank PDFs: all 10 were valid PDFs and all 10 matched the stored SHA-256.
The sanitized per-document and summary reports are included under
\path{provenance/reconstruction_validation/}. SourceShift retains source landing
URLs for its document-level audit trail.
The primary CrossPage OCR path matches the experiment setup:
\begin{verbatim}
python3 data_release/fetch_source_pdfs.py \
  --corpus data_release/release/corpus.jsonl \
  --out-dir /data/nclue/pdfs \
  --resolve-pmc-oa
python3 ../scripts/render_pdf_to_images.py \
  --pdf-dir /data/nclue/pdfs \
  --out-root /data/nclue/pages
mineru -p /data/nclue/pdfs -o /data/nclue/mineru_raw -b pipeline
python3 ../scripts/ocr_pages_mineru.py \
  --mineru-root /data/nclue/mineru_raw \
  --ocr-root /data/nclue/ocr_mineru \
  --corpus data_release/release/corpus.jsonl \
  --write-md
\end{verbatim}
The MinerU adapter reads \path{content_list.json} or \path{middle.json} and
writes one JSON per document page beneath \path{OCR_ROOT}, the same layout used
by the construction pipeline and text retrievers. A Tesseract adapter is also
included as a fallback for reviewers who cannot install MinerU, but it is not the
OCR engine used to produce the reported matrix.

The successful ten-document check directly validates the URL, PDF, and checksum
stage. The commands above continue deterministically through page rendering and
the same MinerU adapter used by construction and retrieval.

\section{Release Files}

\begin{sloppypar}
The Code and Data Supplement is released as a public repository at
\mbox{\url{https://github.com/sunggukcha/n_clue}}. It is self-contained: the
reported numbers are recomputed from files inside the repository, with no network
access, GPU, model download, or non-standard Python package. Its principal files
are:
\end{sloppypar}
\begin{itemize}
\item \texttt{reproduce.py}: standard-library-only one-command regeneration and
claim verification;
\item \texttt{data/v1/} (CrossPage): 1,000 queries, 50,995 qrels, compact 2,021-document
metadata, 70 top-10 rankings, and the 39 query-form rankings;
\item \texttt{data/v2/} (SourceShift): 225-query/500-document robustness set, the initial
8,849-qrel copy, the targeted-revalidation 8,841-qrel copy, and all 39 top-10
rankings;
\item \texttt{reference\_metrics/}: adjudicated paired/clustered outputs for
cross-checking;
\item \texttt{code/}: scoring, query rendering, retrieval, decomposition,
fusion, reranking, reconstruction, and sensitivity source;
\item \texttt{audit/}: the 300-pair audit, expanded-page adjudication, 150-pair
target/control challenge, 75-condition tie-break, administrator records, scorers,
and integrity reports;
\item \texttt{provenance/}: compact OCR-gate, configuration, and blocked-system
records; and
\item \texttt{MANIFEST.json} and \texttt{SHA256SUMS}: inventory and integrity
checks.
\end{itemize}

Two artifacts named in this appendix are withheld from the public repository
because redistributing them would redistribute third-party document content: the
prebuilt annotator HTML packet, which embeds roughly 2,130 near-full-page OCR
excerpts, and the source PDFs, page images, and OCR caches. The packet builders,
the blind audit key, the rubric, the scorers, and all audit \emph{outcomes} are
released, so the packet can be rebuilt locally and the audit re-scored without
it. No reported number reads either artifact. \texttt{THIRD\_PARTY\_NOTICES.md}
in the repository records the per-source terms behind this decision.

From a fresh clone, run:
\begin{verbatim}
python reproduce.py --verify
\end{verbatim}
This recomputes the CrossPage and SourceShift tables, exploratory paired contrasts, exact-family primary
tests, clustered intervals, source-specific deltas, system-rank correlation, and
the targeted revalidation analysis, then checks every file hash. The released
rankings are clipped to rank 10, which is
lossless for the reported @10 rank metrics. The earlier score-margin diagnostic
is not reported because public rank-only files do not retain raw model scores.
The one-command path reproduces the principal score, contrast, and sensitivity
claims from frozen rankings without a GPU. In the repository,
\path{paper_tables/make_capability_strata.py} separately regenerates the failure
decomposition and task-factor table, and fails if any aggregate drifts more than
0.05 points from the published value. The included ranking-generation source
requires the documented OCR/model dependencies and weights for end-to-end
neural regeneration.

%% file: tables/table_primary_scorecard.tex
\begin{tabular}{llp{0.31\textwidth}rrr}
\toprule
Family & System & Multi-condition operation & \textbf{n-Clue Score@10} & Gold Hit@10 & Gold-NDCG@10 \\
\midrule
Lexical & BM25-AND & joint AND query & 26.8 & 57.2 & 32.2 \\
 & BM25 distinct-page & condition lists + distinct pages & 27.9 & 64.7 & 33.3 \\
\addlinespace[1.5pt]
Dense & Jina-v4 joint & one query vector & 17.2 & 51.0 & 21.5 \\
 & Jina-v4 condition-wise & condition lists + harmonic mean & 24.1 & 62.4 & 29.3 \\
 & Qwen3-Emb. 8B joint & one query vector & 19.6 & 50.7 & 22.5 \\
 & Qwen3-Emb. 0.6B condition-wise & condition lists + product & 25.6 & 61.1 & 29.7 \\
\addlinespace[1.5pt]
Visual & ColQwen2.5 & page MaxSim & 22.3 & 62.4 & 29.4 \\
 & ColQwen2.5 top-3 & average three page scores & 25.2 & 71.2 & 34.5 \\
\addlinespace[1.5pt]
Hybrid & BM25 + visual top-3$^{\dagger}$ & equal-weight RRF & \textbf{35.8} & \textbf{81.1} & \textbf{45.6} \\
\addlinespace[1.5pt]
Rerank & BGE-M3 base & fixed text candidates & 12.4 & 39.5 & 14.7 \\
 & BGE reranker & rerank BGE candidates & 9.6 & 21.8 & 8.7 \\
 & Qwen3 text reranker & rerank BGE candidates & 15.0 & 33.0 & 13.7 \\
 & Qwen3-VL 8B reranker & rerank visual candidates & 17.3 & 54.3 & 22.9 \\
\bottomrule
\end{tabular}

%% file: tables/table_failure_decomposition.tex
\begin{tabular}{lrrr}
\toprule
System & Complete first & Subset first & No top-10 gold \\
\midrule
BM25-AND & 26.8 & 30.4 & 42.8 \\
Qwen3-Emb. 8B & 19.6 & 31.1 & 49.3 \\
Qwen3-Emb. 0.6B dec. & 25.6 & 35.5 & 38.9 \\
ColQwen2.5 top-3 & 25.2 & 46.0 & 28.8 \\
Visual + BM25 RRF & 35.8 & 45.3 & 18.9 \\
\bottomrule
\end{tabular}

%% file: tables/table_capability_strata.tex
\begin{tabular}{lrrrrrrr}
\toprule
System & All & $n{=}2$ & $n{=}3$ & $L_{\min}{\leq}30$ & $L_{\min}{>}100$ & $|D^+|{\leq}25$ & $|D^+|{>}50$ \\
 & (1000) & (814) & (186) & (185) & (299) & (146) & (384) \\
\midrule
BM25-AND & 26.8/57.2 & 27.6/57.9 & 23.1/54.3 & 38.4/62.2 & 24.1/51.8 & 36.3/61.6 & 20.6/53.9 \\
Qwen3-Emb. 8B & 19.6/50.7 & 20.9/51.4 & 14.0/47.8 & 18.9/42.7 & 22.7/51.2 & 28.8/63.7 & 13.8/46.1 \\
Qwen3-Emb. 0.6B dec. & 25.6/61.1 & 28.3/61.1 & 14.0/61.3 & 28.6/60.0 & 28.4/63.9 & 34.9/71.9 & 19.5/59.4 \\
ColQwen2.5 top-3 & 25.2/71.2 & 25.1/69.9 & 25.8/76.9 & 29.7/73.5 & 21.1/67.2 & 26.7/75.3 & 22.7/72.1 \\
Visual + BM25 RRF & 35.8/81.1 & 37.0/80.5 & 30.6/83.9 & 41.6/84.9 & 33.1/79.3 & 43.2/83.6 & 31.0/81.8 \\
\bottomrule
\end{tabular}

%% file: tables/table_s1.tex
\begin{landscape}
\begin{center}
\normalsize
\renewcommand{\arraystretch}{1.22}
\setlength{\tabcolsep}{4.5pt}
\begin{tabular}{r p{3.25in} c r c r r}
\hline
Rank & System & Cat. & \textbf{n-Clue Score} & Exact-family 95\% CI & Gold Hit & Gold-NDCG \\
\hline
1 & \texttt{bm25and\_colqwen2.5ptop3} & FUS & 31.1 & [27.4, 35.0] & 74.7 & 40.5 \\
2 & \texttt{bm25decompose\_sdr\_colqwen2.5ptop3} & FUS & 30.0 & [26.5, 33.7] & 77.4 & 40.2 \\
3 & \texttt{bm25and\_qwen3vl2b} & FUS & 29.4 & [25.8, 34.1] & 61.6 & 33.3 \\
4 & \texttt{bm25and\_qwen3emb} & FUS & 29.2 & [25.2, 33.7] & 62.1 & 33.4 \\
5 & \texttt{bm25and\_jinav4} & FUS & 28.9 & [24.9, 33.3] & 67.6 & 35.6 \\
6 & \texttt{bm25and\_bge\_m3} & FUS & 27.8 & [23.7, 32.2] & 60.5 & 31.1 \\
7 & \texttt{bm25and\_e5\_large} & FUS & 27.2 & [23.4, 31.7] & 61.6 & 32.5 \\
8 & \texttt{bm25and\_gte\_large} & FUS & 26.7 & [22.9, 30.9] & 61.8 & 31.2 \\
9 & \texttt{bm25decompose\_sdr\_jinav4} & FUS & 26.5 & [23.2, 29.9] & 72.0 & 34.9 \\
10 & \texttt{bm25dec\_jinav4\_colqwenptop3} & FUS & 26.3 & [23.0, 29.8] & 77.1 & 38.1 \\
11 & \texttt{bm25\_decompose\_sdr} & DEC & 25.3 & [21.8, 28.7] & 66.4 & 33.1 \\
12 & \texttt{qwen3emb0.6b\_dec\_product} & DEC & 25.0 & [22.0, 28.1] & 60.2 & 28.6 \\
13 & \texttt{jinav4\_dec\_hmean} & DEC & 25.0 & [21.7, 28.0] & 64.0 & 30.7 \\
14 & \texttt{bm25\_and\_tuned} & BL & 25.0 & [20.9, 30.0] & 40.0 & 24.5 \\
15 & \texttt{jinav4\_dec\_product} & DEC & 24.9 & [21.5, 28.0] & 63.4 & 30.2 \\
16 & \texttt{e5\_large\_dec\_hmean} & DEC & 24.9 & [21.9, 28.1] & 61.3 & 29.2 \\
17 & \texttt{qwen3emb0.6b\_dec\_hmean} & DEC & 24.7 & [21.8, 27.9] & 58.8 & 28.0 \\
18 & \texttt{e5\_large\_dec\_product} & DEC & 24.2 & [21.1, 27.3] & 61.2 & 29.0 \\
19 & \texttt{bm25\_and} & BL & 23.8 & [20.1, 28.5] & 38.5 & 23.4 \\
20 & \texttt{bm25\_decompose\_hmean} & DEC & 23.3 & [19.8, 27.4] & 52.7 & 24.9 \\
21 & \texttt{bm25or\_jinav4} & FUS & 23.3 & [19.9, 27.0] & 65.4 & 30.4 \\
22 & \texttt{e5\_large\_dec\_min} & DEC & 22.8 & [19.5, 26.3] & 51.4 & 23.3 \\
\hline
\end{tabular}
\par\smallskip\textbf{Table S1a.} CrossPage matrix ranked by n-Clue Score, ranks 1--22. Gold Hit and Gold-NDCG are supporting subscores.
\end{center}
\end{landscape}

\begin{landscape}
\begin{center}
\normalsize
\renewcommand{\arraystretch}{1.22}
\setlength{\tabcolsep}{4.5pt}
\begin{tabular}{r p{3.25in} c r c r r}
\hline
Rank & System & Cat. & \textbf{n-Clue Score} & Exact-family 95\% CI & Gold Hit & Gold-NDCG \\
\hline
23 & \texttt{bm25\_decompose\_product} & DEC & 22.5 & [19.0, 26.4] & 55.0 & 25.9 \\
24 & \texttt{jinav4\_dec\_min} & DEC & 22.4 & [19.1, 25.9] & 49.0 & 23.8 \\
25 & \texttt{colqwen2.5\_ptop3} & RET & 22.4 & [19.4, 25.5] & 68.2 & 31.6 \\
26 & \texttt{jinav4\_colqwen2.5} & FUS & 21.9 & [18.9, 25.1] & 66.9 & 30.6 \\
27 & \texttt{colqwen2.5\_ptop2} & RET & 21.8 & [18.4, 25.2] & 65.9 & 30.3 \\
28 & \texttt{jinav4\_colqwen2.5ptop3} & FUS & 21.5 & [18.4, 24.7] & 69.6 & 31.3 \\
29 & \texttt{gte\_large\_dec\_product} & DEC & 21.2 & [18.4, 24.0] & 54.9 & 23.7 \\
30 & \texttt{gte\_large\_dec\_hmean} & DEC & 21.2 & [18.4, 24.2] & 54.9 & 23.8 \\
31 & \texttt{colqwen2.5} & RET & 21.2 & [18.5, 23.9] & 59.2 & 28.3 \\
32 & \texttt{bm25\_or} & BL & 20.6 & [17.5, 24.2] & 53.1 & 24.1 \\
33 & \texttt{bm25\_condition\_text} & BL & 20.6 & [17.5, 24.2] & 53.1 & 24.1 \\
34 & \texttt{qwen3emb0.6b\_dec\_min} & DEC & 20.2 & [17.3, 23.5] & 44.0 & 20.9 \\
35 & \texttt{bge\_m3\_dec\_hmean} & DEC & 19.9 & [16.8, 23.3] & 46.6 & 19.8 \\
36 & \texttt{rerank\_qwen3vl2b} & RER & 19.7 & [16.6, 22.9] & 58.7 & 25.4 \\
37 & \texttt{bm25\_decompose\_min} & DEC & 19.4 & [16.2, 22.9] & 45.3 & 19.9 \\
38 & \texttt{jinav4\_qwen3vl8b} & FUS & 19.3 & [16.0, 22.2] & 59.5 & 25.3 \\
39 & \texttt{bge\_m3\_dec\_product} & DEC & 19.2 & [16.2, 22.6] & 46.1 & 19.4 \\
40 & \texttt{qwen3vl8b} & RET & 19.0 & [15.7, 22.4] & 52.1 & 21.7 \\
41 & \texttt{gte\_large\_dec\_min} & DEC & 18.9 & [16.3, 21.7] & 41.6 & 18.2 \\
42 & \texttt{jinav4\_paraphrased} & QF & 18.5 & [15.5, 21.6] & 49.7 & 21.0 \\
43 & \texttt{text\_jinav4} & RET & 18.2 & [15.2, 21.4] & 54.5 & 22.8 \\
44 & \texttt{jinav4\_qwen3vl2b} & FUS & 18.1 & [15.4, 20.7] & 54.8 & 24.1 \\
\hline
\end{tabular}
\par\smallskip\textbf{Table S1b.} CrossPage matrix ranked by n-Clue Score, ranks 23--44. Gold Hit and Gold-NDCG are supporting subscores.
\end{center}
\end{landscape}

\begin{landscape}
\begin{center}
\normalsize
\renewcommand{\arraystretch}{1.22}
\setlength{\tabcolsep}{4.5pt}
\begin{tabular}{r p{3.25in} c r c r r}
\hline
Rank & System & Cat. & \textbf{n-Clue Score} & Exact-family 95\% CI & Gold Hit & Gold-NDCG \\
\hline
45 & \texttt{e5\_large\_paraphrased} & QF & 17.8 & [15.1, 20.8] & 46.4 & 19.9 \\
46 & \texttt{text\_qwen3emb0.6b} & RET & 17.7 & [15.0, 20.7] & 48.4 & 20.6 \\
47 & \texttt{text\_qwen3emb8b} & RET & 17.7 & [14.8, 20.8] & 49.6 & 21.0 \\
48 & \texttt{text\_e5\_large} & RET & 17.2 & [14.3, 20.3] & 44.9 & 19.4 \\
49 & \texttt{text\_qwen3emb4b} & RET & 17.1 & [14.1, 20.2] & 47.9 & 20.6 \\
50 & \texttt{qwen3vl2b} & RET & 17.1 & [14.2, 19.9] & 44.3 & 19.2 \\
51 & \texttt{colbertv2.0} & RET & 17.0 & [14.3, 19.8] & 53.8 & 22.5 \\
52 & \texttt{qwen3emb\_paraphrased} & QF & 16.7 & [13.9, 19.5] & 46.8 & 19.6 \\
53 & \texttt{dse\_qwen2\_2b} & RET & 16.4 & [13.8, 19.2] & 39.0 & 14.9 \\
54 & \texttt{bge\_m3\_dec\_min} & DEC & 16.1 & [13.6, 18.7] & 38.8 & 16.0 \\
55 & \texttt{bge\_m3\_sparse} & RET & 15.8 & [12.9, 19.2] & 49.5 & 20.8 \\
56 & \texttt{bge\_m3\_mv} & RET & 15.3 & [13.0, 18.0] & 46.4 & 18.4 \\
57 & \texttt{rerank\_qwen3vl8b} & RER & 15.1 & [11.8, 18.7] & 52.4 & 21.6 \\
58 & \texttt{bm25\_page\_aggregated} & BL & 14.5 & [11.8, 17.2] & 41.8 & 16.9 \\
59 & \texttt{frequency\_prior} & BL & 14.1 & [11.5, 17.2] & 40.6 & 14.6 \\
60 & \texttt{text\_gte\_large} & RET & 13.5 & [10.9, 16.3] & 43.7 & 16.4 \\
61 & \texttt{text\_bge\_m3} & RET & 13.3 & [10.8, 16.0] & 40.8 & 15.3 \\
62 & \texttt{gte\_large\_paraphrased} & QF & 13.3 & [10.8, 15.9] & 44.7 & 16.8 \\
63 & \texttt{rerank\_qwen3r8b} & RER & 13.2 & [10.5, 16.0] & 29.8 & 11.8 \\
64 & \texttt{bge\_m3\_paraphrased} & QF & 12.8 & [10.5, 15.4] & 40.3 & 14.8 \\
65 & \texttt{rerank\_bge\_m3} & RER & 9.4 & [7.4, 11.5] & 24.6 & 9.4 \\
66 & \texttt{random} & BL & 0.4 & [0.1, 0.8] & 0.5 & 0.1 \\
\hline
\end{tabular}
\par\smallskip\textbf{Table S1c.} CrossPage matrix ranked by n-Clue Score, ranks 45--66. Gold Hit and Gold-NDCG are supporting subscores.
\end{center}
\end{landscape}

\begin{center}
\small
\setlength{\tabcolsep}{4.0pt}
\begin{tabular}{p{1.8in} p{2.2in} r r r}
\hline
Control & Privileged input & n-Clue Score & Gold Hit & Gold-NDCG \\
\hline
\texttt{surface\_form\_and} & all-alias construction facts & 100.0 & 100.0 & 99.5 \\
\texttt{surface\_form\_count} & gold-defining fact-count oracle & 100.0 & 100.0 & 99.7 \\
\texttt{exact\_name\_and} & canonical construction facts & 16.7 & 16.7 & 9.6 \\
\texttt{single\_page\_surface\_and} & privileged single-page floor & 0.0 & 0.0 & 0.0 \\
\hline
\end{tabular}
\par\smallskip\textbf{Table S1d.} Unranked construction-fact controls. Every row reads the immutable construction facts and aliases rather than only released query text and OCR, so none is a fair retriever. The 100-point controls remain consistent after repair because every query retains a construction gold at the first eligible rank; repaired subsets can only cease to block that gold.
\end{center}

%% file: tables/table_headline_sensitivity.tex
\begin{center}
\small
\renewcommand{\arraystretch}{1.08}
\setlength{\tabcolsep}{2.6pt}
\begin{tabular}{lrrrrr}
\hline
Contrast & $\Delta$Score & Exact-family 95\% CI & $\Delta$Hit & $\Delta$NDCG & Sel.+Holm $p$ \\
\hline
Scale & +0.0 & [-2.72, +2.75] & +1.2 & +0.4 & 1.000 \\
Page aggregation & +1.2 & [-1.63, +4.00] & +9.0 & +3.3 & 1.000 \\
Lexical composition & +1.5 & [-3.08, +5.91] & +27.9 & +9.8 & 1.000 \\
Jina composition & +6.8 & [+3.70, +9.89] & +9.5 & +7.9 & 0.005 \\
Qwen composition & +7.3 & [+3.86, +10.58] & +11.8 & +8.0 & 0.002 \\
Complementarity & +8.7 & [+5.63, +11.94] & +6.5 & +8.9 & 0.005 \\
Text reranking & -3.9 & [-7.14, -0.71] & -16.2 & -5.9 & 0.168 \\
Visual reranking & -6.1 & [-9.56, -2.79] & -6.8 & -6.7 & 0.008 \\
\hline
\end{tabular}
\par\smallskip\textbf{Table S1e.} Primary-metric inference for all eight displayed CrossPage contrasts. Effects and intervals are percentage points. Endpoint selection and multiplicity are tested on n-Clue Score, the declared primary metric; NDCG is a subscore.
\end{center}

\begin{center}
\small
\renewcommand{\arraystretch}{1.08}
\setlength{\tabcolsep}{3.0pt}
\begin{tabular}{lrrrr}
\hline
Contrast & Common $n$ & $\Delta D$ & Base M/P/G & Variant M/P/G \\
\hline
Scale & 378 & -0.011 & 0.516/0.307/0.177 & 0.504/0.319/0.177 \\
Page aggregation & 545 & +0.011 & 0.408/0.380/0.212 & 0.318/0.458/0.224 \\
Lexical composition & 307 & -0.212 & 0.615/0.147/0.238 & 0.336/0.411/0.253 \\
Jina composition & 470 & +0.081 & 0.455/0.363/0.182 & 0.360/0.390/0.250 \\
Qwen composition & 414 & +0.075 & 0.516/0.307/0.177 & 0.398/0.352/0.250 \\
Complementarity & 660 & +0.103 & 0.318/0.458/0.224 & 0.253/0.436/0.311 \\
Text reranking & 143 & -0.007 & 0.592/0.275/0.133 & 0.754/0.152/0.094 \\
Visual reranking & 444 & -0.077 & 0.408/0.380/0.212 & 0.476/0.373/0.151 \\
\hline
\end{tabular}
\par\smallskip\textbf{Table S1f.} Common-support ordering and unconditional outcomes. $\Delta D$ is variant minus base in $1-P(\mathrm{subset>gold})$ where both retrieve a gold. M/P/G are unconditional miss, subset-first, and gold-first proportions.
\end{center}
\clearpage

\begin{center}
\small
\renewcommand{\arraystretch}{1.08}
\setlength{\tabcolsep}{2.2pt}
\begin{tabular}{lrrrr}
\hline
Contrast & Document macro & Crossed all-judged 95\% CI & Leave-one-doc range & One-query/doc 95\% CI \\
\hline
Scale & +2.1 & [-2.58, +2.56] & [-1.77, +1.55] & [+0.35, +3.69] \\
Page aggregation & -3.3 & [-1.40, +3.80] & [-0.76, +4.03] & [-5.35, -1.32] \\
Lexical composition & +0.6 & [-2.00, +4.98] & [-3.67, +5.90] & [-1.74, +2.99] \\
Jina composition & +5.7 & [+3.88, +9.73] & [+5.11, +10.70] & [+3.62, +7.72] \\
Qwen composition & +7.6 & [+4.37, +10.35] & [+5.43, +9.43] & [+5.63, +9.60] \\
Complementarity & +7.0 & [+6.07, +11.43] & [+6.64, +11.03] & [+5.15, +8.90] \\
Text reranking & -3.8 & [-6.66, -1.18] & [-6.39, -2.19] & [-5.49, -2.02] \\
Visual reranking & -3.7 & [-9.09, -3.19] & [-8.31, -4.63] & [-5.84, -1.46] \\
\hline
\end{tabular}
\par\smallskip\textbf{Table S1g.} Shared-document sensitivity for n-Clue Score. All values are percentage points. The four weighting/deletion schemes expose document reuse; none asserts independent query sampling.
\end{center}

\begin{center}
\small
\renewcommand{\arraystretch}{1.08}
\setlength{\tabcolsep}{4.0pt}
\begin{tabular}{lrrrr}
\hline
Contrast & Observed & $\leq1$ & $\leq2$ & $\leq3$ \\
\hline
Scale & +0.0 & -48.6 & -48.6 & -48.6 \\
Page aggregation & +1.2 & -44.4 & -44.4 & -44.4 \\
Lexical composition & +1.5 & -23.1 & -23.1 & -23.1 \\
Jina composition & +6.8 & -51.4 & -51.4 & -51.4 \\
Qwen composition & +7.3 & -49.6 & -49.6 & -49.6 \\
Complementarity & +8.7 & -24.6 & -24.6 & -24.6 \\
Text reranking & -3.9 & -78.1 & -78.1 & -78.1 \\
Visual reranking & -6.1 & -69.0 & -69.0 & -69.0 \\
\hline
\end{tabular}
\par\smallskip\textbf{Table S1h.} Claim-specific adversarial qrel bounds in n-Clue Score points. Each column minimizes variant-minus-base score after promoting up to $k$ currently judged top-10 subsets per query. This is an exact bound, not an error model; qrel-absent documents remain outside its scope.
\end{center}

%% file: supplement_audit.tex
\section{Evidence Audit and Adjudication}

This section documents the label-validation process for the n-Clue measurement
instrument after construction. It presents the evaluator-facing HTML, expanded-page adjudication,
the pre-specified evidence-asymmetric policy, and a second 125-target
retrieval-conditioned qrel challenge with 25 blinded controls. Agreement,
response integrity, exact qrel transitions, and evaluation effects are reported
for both stages. Evaluators never needed to read a whole document: the initial
pass showed three condition-specific page contexts and adjudication/challenge
polls showed six to nine pages from the start. Exact quotes, page selections, and
distinct-page feasibility were checked mechanically.

\subsection{Scope and Sampling}

The audit unit is a query--document pair, but judgments are made independently
for every query condition. The initial stratified packet contains 300 pairs and 710
condition decisions: 100 pairs from CrossPage and 200 from the four sources
retained by SourceShift. Stable query/document identifiers, benchmark
grades, gold/partial labels, construction pages, and retrieval routes were kept
in a separate administrator key. The later robustness packet is separate: all
125 SourceShift subsets capable of reducing the frozen fusion--BM25 top-10 contrast
are targets, and 25 neutral controls are interleaved. It contains 150 pairs and
320 condition decisions; target/control status and the contrast itself are hidden.

\begin{center}
\small
\setlength{\tabcolsep}{3.1pt}
\begin{tabular}{llrrr}
\toprule
Release & Source & Pairs & Pre-gold & Pre-subset \\
\midrule
CrossPage & World Bank & 98 & 50 & 48 \\
CrossPage & PMC OA & 2 & 0 & 2 \\
\midrule
SourceShift & arXiv & 50 & 25 & 25 \\
SourceShift & ERIC & 50 & 25 & 25 \\
SourceShift & Federal Register & 50 & 25 & 25 \\
SourceShift & PMC OA & 50 & 22 & 28 \\
\midrule
All & & 300 & 147 & 153 \\
\bottomrule
\end{tabular}
\captionof{table}{Stratified audit sample. Counts describe the sampled packet, not the
population distribution of qrels.}
\label{tab:sampling}
\end{center}

The audit asks whether displayed text is \emph{necessary evidence}, not whether
it is merely related. Every condition supplies (i) a complete English
requirement, (ii) explicit ``counts'' criteria, (iii) explicit ``does not count''
criteria, and where needed (iv) a plain-language note. External professional
knowledge is prohibited. A positive answer requires one or more page selections
and an exact supporting quote; a negative answer means only that no displayed
passage establishes the requirement.

The task targeted English-proficient generalists: rubrics defined needed domain
terms and disallowed outside inference. Recruitment, professional-background,
and compensation metadata were not collected, so we make no expertise-stratified
or timing-based claims. No demographic or sensitive personal data were requested.

\begin{figure*}[t]
\centering
\includegraphics[width=0.84\textwidth]{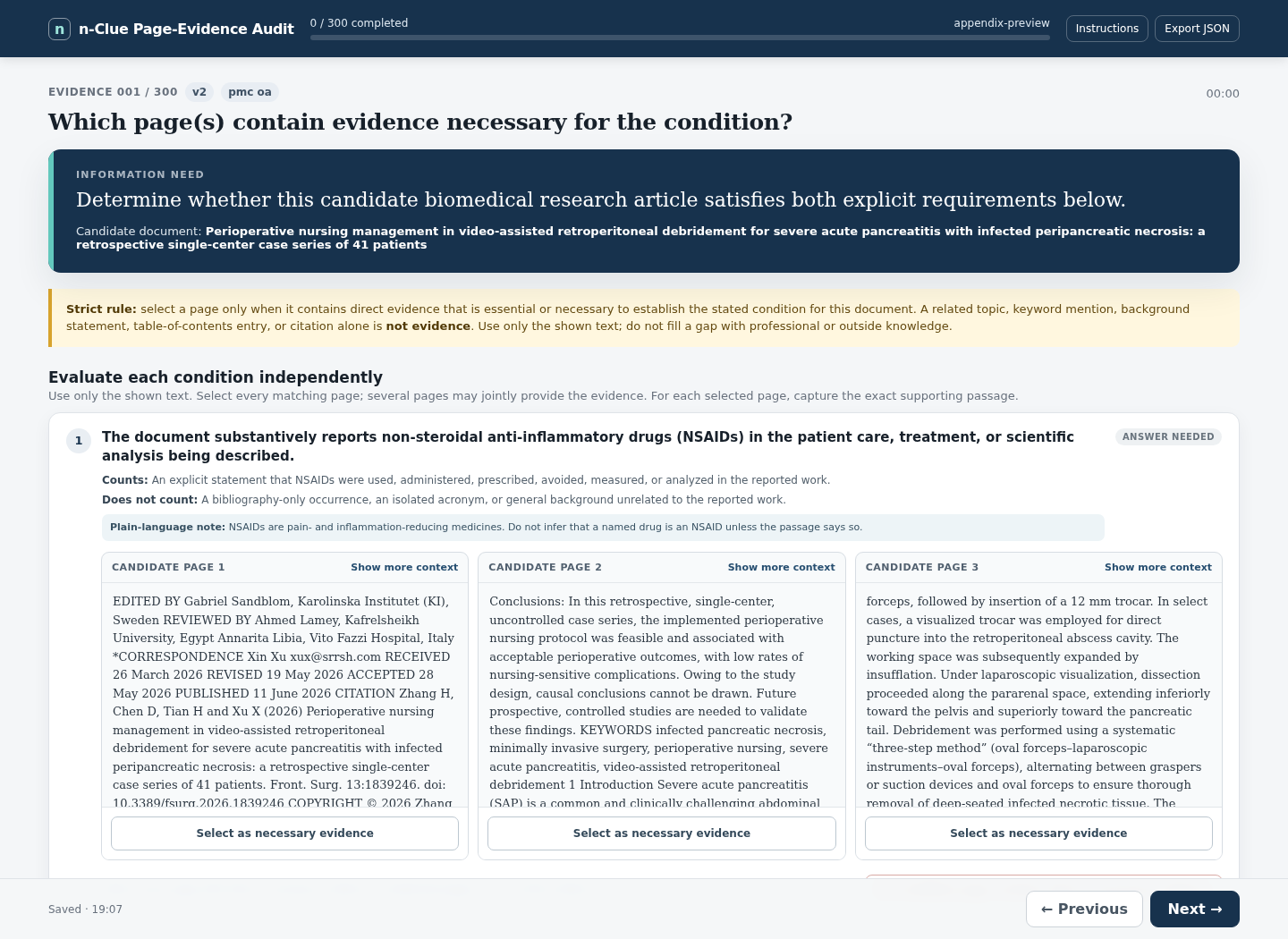}
\caption{The evaluator-facing SourceShift evidence-audit interface. The complete
information need, strict necessary-evidence rule, condition-specific rubric, and
all three candidate pages are visible together. Positive decisions require page
selection and a verbatim quote; ``related'' and ``uncertain'' are not response
options. The distributed interface is entirely English.}
\label{fig:audit-ui}
\end{figure*}

\subsection{Initial Blind Audit}

\subsubsection{Interface and validation gates}

The self-contained HTML stores all text locally and requires no network access or
PDF reader. It supports autosave, import/resume, keyboard navigation, always-live
Previous/Next controls, and final JSON export. Before accepting an item, the
browser checks that every condition has a binary answer; every positive answer
has at least one in-pool page and a nontrivial quote; every quote is a normalized
substring of its selected page; and every negative answer has no page or quote.
The released packet contains 300 pairs, 710 rubrics, and 2,130 nonempty page
contexts, with no private-field leaks or broad fallback rubrics.

Two independent evaluators completed every pair and condition. Evaluator A
selected direct evidence for 560 conditions and evaluator E for 555. Their
exports have no structural, page-pool, or quote-matching errors.

\begin{center}
\small
\begin{tabular}{lr}
\toprule
Initial-audit statistic & Result \\
\midrule
Completed pairs / conditions & 300 / 710 \\
Exact derived-grade agreement & 254/300 (84.7\%) \\
Linearly weighted grade $\kappa$ & .801 \\
Condition support agreement & 661/710 (93.1\%) \\
Exact condition page-set agreement & 78.7\% \\
Queries judged clear by both & 300/300 \\
Selected quotes passing exact check & 2,304/2,304 \\
\bottomrule
\end{tabular}
\captionof{table}{Agreement and integrity in the initial 300-pair blind audit.}
\label{tab:initial-agreement}
\end{center}

Exact grade agreement has a Wilson 95\% interval of [80.2, 88.3]\%; condition
agreement has interval [91.0, 94.7]\% and binary Cohen's $\kappa=.795$. Exact
grade agreement is 85.3\% for $n=2$ and 83.6\% for $n=3$, and 81.0\% on
CrossPage versus 86.5\% on SourceShift. The pooled linearly weighted grade
$\kappa$ is therefore a
compact summary across two grade scales, not a substitute for these strata.

\begin{center}
\small
\begin{tabular}{lrrrr}
\toprule
& \multicolumn{4}{c}{Evaluator E grade} \\
Evaluator A grade & 0 & 1 & 2 & 3 \\
\midrule
0 & 2 & 6 & 0 & 0 \\
1 & 2 & 72 & 14 & 0 \\
2 & 0 & 11 & 128 & 1 \\
3 & 0 & 1 & 11 & 52 \\
\bottomrule
\end{tabular}
\captionof{table}{Initial derived-grade confusion matrix. Rows sum to A's decisions and
columns to E's; grade 3 occurs only for three-condition pairs.}
\label{tab:grade-confusion}
\end{center}

\subsubsection{Evidence-asymmetric repair policy}

Presence and absence do not have equal evidential status. Two independent valid
quotes can establish that a supposed partial satisfies an additional condition;
failure to find a quote in three pages cannot establish document-wide absence.
We fixed the following policy before receiving the expanded-page responses:

\begin{enumerate}
\item Accept A+E consensus increases supported by valid direct quotes and a
distinct-page assignment. This resolves 50 positive increases immediately.
\item Send every A/E grade split (46 pairs) to expanded-page adjudication.
\item Send every A+E consensus decrease (29 pairs) to expanded-page
adjudication rather than privileging the old qrel or treating three-page absence
as final.
\end{enumerate}

The resulting queue has 75 unique pairs and 94 unresolved conditions: 47 A/E
splits and 47 conditions both evaluators marked negative. Another 86 conditions
within those pairs were already directly supported and remain locked.

\subsection{Expanded-Page Adjudication Poll}

\begin{figure*}[t]
\centering
\includegraphics[width=0.84\textwidth]{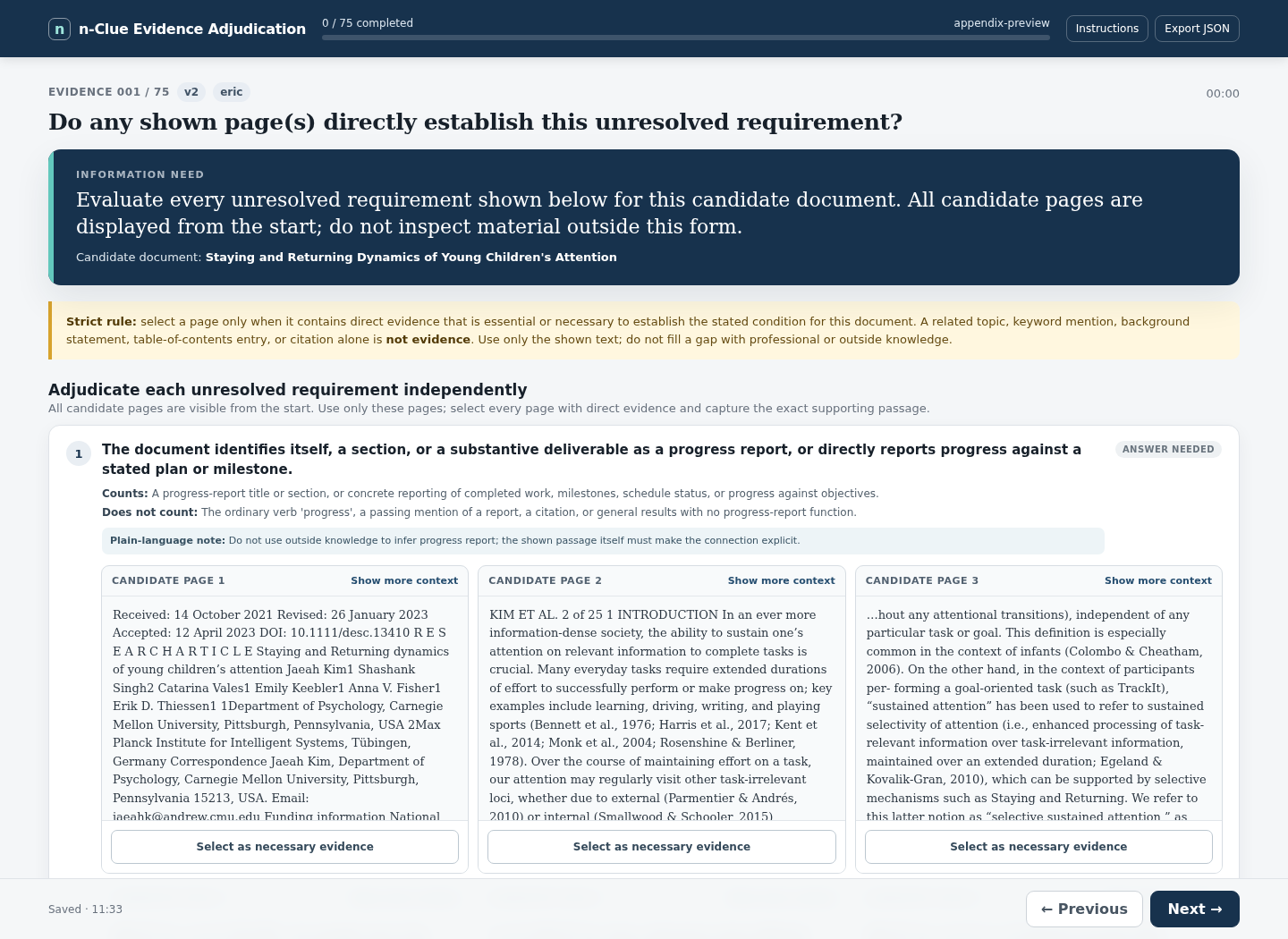}
\caption{Expanded-page adjudication poll. All six to nine pages are visible from
the start; there is no hidden rescue stage. The evaluator sees neither the prior
qrel nor A/E votes, candidate ranks, or retrieval routes. The same binary
necessary-evidence rule and exact-quote validation apply.}
\label{fig:adj-ui}
\end{figure*}

\subsubsection{High-recall candidate pool}

Each unresolved condition receives a deduplicated union of original candidate
pages, pages already selected by evaluators, exact/alias lexical hits, BM25 page
retrieval, OCR-tolerant fuzzy retrieval, and construction evidence when
available. Route and rank are hidden in the evaluator form. Candidate membership
counts in Table~\ref{tab:pool} are nonexclusive because one page can be recovered
by several routes.

\begin{center}
\small
\begin{tabular}{lr}
\toprule
Candidate-pool statistic & Count \\
\midrule
Pairs / unresolved conditions & 75 / 94 \\
Displayed page contexts & 830 \\
Pages per condition (min / mean / max) & 6 / 8.83 / 9 \\
Prior evaluator-selected memberships & 92 \\
Original three-page memberships & 282 \\
Exact/alias lexical memberships & 321 \\
BM25 memberships & 344 \\
OCR-fuzzy memberships & 349 \\
Construction-fact memberships & 98 \\
\bottomrule
\end{tabular}
\captionof{table}{Expanded-page pool composition. Membership rows are not mutually
exclusive.}
\label{tab:pool}
\end{center}

As a held-out check, the pool is evaluated on the 86 locked positive conditions
in the same queued documents. It recovers evidence for 86/86 by rank six and
86/86 by rank nine; all previously submitted positive pages are present by rank
nine. This check establishes recall for known positives under the audit
distribution, not exhaustive document-level recall for unknown negatives.

\subsubsection{Adjudication agreement and integrity}

Three adjudication responses complete all 75 pairs and 94 unresolved conditions.
They agree unanimously on 80/94 binary decisions (85.1\%); Fleiss $\kappa=.636$.
All 613 selected-page quotes are found exactly in the displayed text, and all
selected pages belong to the relevant pool. Pairwise condition agreement is
86.2\%, 92.6\%, and 91.5\%; the lower chance-corrected agreement reflects the
high positive prevalence in this deliberately difficult queue.
Responses A and E come from the two initial evaluators; F is the additional
response. Thus this agreement measures repeated adjudication evidence, not three
fresh independent annotators, and timing metadata are not used as provenance.

\begin{center}
\small
\setlength{\tabcolsep}{4pt}
\begin{tabular}{lrrrr}
\toprule
Response & Positive & Negative & Pages & Quotes exact \\
\midrule
A & 76 & 18 & 201 & 201/201 \\
E & 77 & 17 & 214 & 214/214 \\
F & 83 & 11 & 198 & 198/198 \\
\midrule
All & 236 & 46 & 613 & 613/613 \\
\bottomrule
\end{tabular}
\captionof{table}{Adjudication response integrity. Timing fields are intentionally not
used in any validity or agreement calculation.}
\label{tab:adj-integrity}
\end{center}

\subsection{Evidence-Aware Final Decisions}

A raw two-of-three vote cannot enforce distinct evidence pages or correct a
clear miss visible in displayed text. An administrative check of all 14
non-unanimous conditions applied the same rubric and evidence; it is neither a
fourth vote nor included in $\kappa$. Three decisions depart from raw majority:

\begin{itemize}
\item \textbf{J011:} a displayed page has a substantive ``Conclusions'' heading
and body; the majority-negative condition is corrected to positive.
\item \textbf{J060:} displayed regulatory pages explicitly require analysis of
both costs and benefits; the majority-negative condition is corrected to
positive.
\item \textbf{J061:} both semantic conditions are supported, but only on the same
page. The Hall matching gate fails, so the document remains partial rather than
being promoted to gold.
\end{itemize}

No other majority-gold case fails the distinct-page check. The final manifest
changes 65/300 grades: 62 partial-to-gold, two gold-to-partial, and one internal
partial increase. Table~\ref{tab:transitions} gives the exact transitions; one
$1\rightarrow2$ case remains partial because it has $n=3$.

\begin{center}
\small
\begin{tabular}{lrr}
\toprule
Grade transition & Count & Interpretation \\
\midrule
$1\rightarrow2$ & 37 & 36 to gold; 1 remains partial \\
$2\rightarrow3$ & 26 & 26 to gold \\
$2\rightarrow1$ & 2 & 2 gold to partial \\
\midrule
All changes & 65 & net +60 sampled gold \\
\bottomrule
\end{tabular}
\captionof{table}{Final grade transitions after initial consensus repair and expanded-page
adjudication.}
\label{tab:transitions}
\end{center}

\begin{center}
\small
\begin{tabular}{llrr}
\toprule
Release & Source & Sampled & Corrected \\
\midrule
CrossPage & All (98 WB, 2 PMC) & 100 & 20 \\
\midrule
SourceShift & arXiv & 50 & 13 \\
SourceShift & ERIC & 50 & 8 \\
SourceShift & Federal Register & 50 & 13 \\
SourceShift & PMC OA & 50 & 11 \\
\midrule
All & & 300 & 65 \\
\bottomrule
\end{tabular}
\captionof{table}{Corrections by audit stratum. These rates must not be reweighted into a
population error estimate.}
\label{tab:source-corrections}
\end{center}

\subsection{Targeted Residual-Qrel Challenge}

The stratified audit was not designed to find every false negative in a system's
top ranks. We therefore formed a separate robustness set containing every
already-judged SourceShift subset in the frozen fusion/BM25 top-10 union whose
promotion could reduce that contrast: 125 targets. Twenty-five neutral controls
were interleaved. System, rank, source, original grade, and target/control status
were hidden, and all 150 pairs used the same binary necessary-evidence rule and
six-to-nine-page display as the expanded-page poll.

Integrity review found two nominal first-pass files identical at all 320
condition decisions, selected pages, and quotes; they are collapsed to one
response stream rather than counted twice. The remaining unique stream agrees on
245/320 conditions. A fresh evaluator completed a 75-condition tie-break over 61
pairs; all 136 selected quotes exactly match displayed text. The tie-breaker
selects the duplicate-collapsed stream on 22 conditions and the other stream on
53. One would-be gold fails the distinct-page gate. Final target transitions are
$1\!\rightarrow\!0$ (8), $1\!\rightarrow\!1$ (23),
$1\!\rightarrow\!2$ (82), $1\!\rightarrow\!3$ (6),
$2\!\rightarrow\!2$ (1), and $2\!\rightarrow\!3$ (5): 101/125 grades change,
90 targets become gold, and 23/25 control grades are retained.

\subsection{Effect on Evaluation}

Table~\ref{tab:prepost} separates the two validation stages. The stratified audit
updates 20 CrossPage and 45 SourceShift qrels; one demotion removes a zero-gold
SourceShift query. Across all 70 CrossPage systems, pre/post rank correlation is $.9998$, representative
gold-only NDCG@10 moves by at most $.0012$, and the largest movement is one
position. The later challenge changes no CrossPage labels. On SourceShift it removes eight
zero-grade target rows and changes the frozen raw contrast: both system scores
increase, but BM25-AND increases more, shrinking the gap to $+.016$ with
query-paired $p=.381$ and shared-gold-family $p=.402$.

\begin{center}
\scriptsize
\setlength{\tabcolsep}{1.5pt}
\begin{tabular}{@{}lrrr@{}}
\toprule
Quantity & Construction & Strat. audit & Target challenge \\
\midrule
Reviewed grades changed & -- & 65/300 & 101/125 \\
SourceShift queries & 226 & 225 & 225 \\
SourceShift nonzero qrels & 8,858 & 8,849 & 8,841 \\
\midrule
CrossPage fusion & .406 & .405 & .405 \\
CrossPage decompose-SDR & .332 & .331 & .331 \\
CrossPage BM25-AND & .234 & .234 & .234 \\
\midrule
SourceShift raw fusion & .293 & .284 & .322 \\
SourceShift raw BM25-AND & .183 & .177 & .305 \\
SourceShift fusion$-$BM25 gap & +.110 & +.107 & +.016 \\
SourceShift paired test & $p=.0001$ & $p=.0001$ & $p=.381$ \\
\bottomrule
\end{tabular}
\captionof{table}{Evaluation impact of both validation stages. Values are
gold-only NDCG@10 unless otherwise indicated. The target challenge is deliberately
contrast-conditioned and therefore tests robustness rather than estimating an
unbiased leaderboard.}
\label{tab:prepost}
\end{center}

The final raw crossed query/gold-document interval is $[-.036,.074]$ and the
crossed query/all-judged-document interval is $[-.021,.055]$. Source deltas are
heterogeneous: $+.128$ arXiv, $+.059$ ERIC, $-.043$ Federal Register, and $+.072$
PMC OA. These numbers are valid for the reviewed qrel copy, but target selection
precludes treating its correction rate or system ordering as population-unbiased.